\documentclass{hld2026} 

\graphicspath{ {figures/} }

\usepackage{graphicx}
\usepackage{subcaption}
\usepackage{booktabs} 
\usepackage{amsmath,amssymb} 
\usepackage{mathtools}

\title[Optimizer-Dependent Hessian Dynamics]{Characterizing Optimizer-Dependent Training Dynamics Through Hessian Eigenvector Displacement and Localization}

\hldauthor{%
\Name{Marcelina Marjankowska}\Email{marcelina.marjankowska.24@ucl.ac.uk}\\
\addr {Financial Computing and Analytics Group, University College London} 
\AND
\Name{Valerio Modugno} \Email{v.modugno@ucl.ac.uk}\\
\addr {Robot Perception and Learning Lab, University College London} 
\AND
\Name{Paolo Barucca} \Email{p.barucca@ucl.ac.uk}\\
\addr {Financial Computing and Analytics Group, University College London}}

\begin{document}

\maketitle

\begin{abstract}%
Hessian spectral properties are a standard tool in analysing neural-network training, with eigenvalues linked to sharpness, generalization, and optimization dynamics \cite{baskerville_universal_2022, yao2018hessian, pmlr-v97-ghorbani19b}. Eigenvalues quantify curvature magnitude, while eigenvectors identify \emph{which parameters} generate that curvature. In this work, we study how the leading Hessian \emph{eigenvectors} evolve during training and how they affect the learning trajectories. 
We track the training dynamics of multilayer perceptrons on a classification problem and measure eigenvector dynamics through two complementary statistics: (i) displacement over time, inspired by analyses of glassy systems \cite{baity-jesi_comparing_2018}, and (ii) localization via the inverse participation ratio. The metrics are compared against a random null model of the Hessian induced by the architecture.
Our results reveal clear optimizer-dependent behaviour. SGD leads to progressively more stable leading curvature directions, while Adam exhibits substantially stronger reorganization of eigenvectors throughout training. We also observe a localization phenomenon under Adam, where a small subset of parameters contributes disproportionately to the leading curvature directions. These results suggest that Hessian eigenvector dynamics capture key differences in optimizer behaviour and the resulting training trajectories.

\end{abstract}


\section{Introduction}
Understanding neural-network training remains challenging because optimization takes place in a high-dimensional and highly non-convex loss landscape. One useful descriptor of the local loss is the Hessian, i.e. the matrix of second derivatives of the loss with respect to model parameters.
Characterising local curvature via the Hessian is central for understanding sensitivity to parameter perturbations and for studying properties of the training dynamics, including generalization-relevant behaviour and approximate Bayesian interpretations of uncertainty.
Local curvature, tracked via the Hessian spectral properties, can be used to study sharpness, flatness, and ruggedness, and its evolution can be tracked not only at convergence but also during the training dynamics. 
Multiple works investigate the Hessian eigenvalues connecting them to generalization \cite{foret_sharpness-aware_2021}, robustness \cite{yao2018hessian} and training dynamics \cite{pmlr-v97-ghorbani19b, baskerville_universal_2022, alain_negative_2019}. 
A natural next question is whether the corresponding eigenvectors, which additionally encode the \emph{directions} of curvature in parameter space, contain useful information beyond the spectrum alone.
We ask the following questions: 
\begin{itemize}
    \item During mini-batch training, does leading Hessian eigenvector stabilize, drift gradually, or reorganize entirely?
    \item Do different optimizers traverse similar geometric trajectories, or do they induce qualitatively distinct forms of landscape exploration?
\end{itemize} 
By using the eigenvector displacement and localization metrics, we find that optimizer choice clearly affects the dominant curvature variability and results in exploration of qualitatively different landscape regions. 

Prior work shows that the \emph{Hessian is approximately low-rank}: in a $k$-class classification problem, the top $k$ eigenvalues are separated from the bulk. Moreover, the leading top-$k$ Hessian subspace remains relatively stable over time \cite{bao_hessian_inertia_2023, gur-ari_gradient_2018}. 
Stability of a subspace does not imply stability of its individual eigenvectors: directions within the top eigenspace may rotate or swap order during training. Most prior work therefore uses subspace-level metrics \cite{gur-ari_gradient_2018, andreyev_edge_2025}. Here, instead, we explicitly track the evolution of individual leading eigenvectors.
This distinction is potentially important: recent work suggests that eigenvector instabilities may help optimization explore new regions of the loss landscape \cite{WANG2026108874}. 
Further, tracking the Hessian spectral dynamics can capture the essential detectability transitions that may prevent learning in datasets with low signal-to-noise ratios \cite{bonnaire2025role}. 
By focusing on mini-batch training and comparing optimizers, we extend prior Hessian-based analyses to more realistic modern training regimes.

\section{Quantifying Eigenvector Dynamics}

Let $\theta \in \mathbb{R}^N$ denote the model parameters, and let $\mathcal{L}(X,Y;\theta)$ be the loss function defined on a dataset $(X,Y)$. Training proceeds via an optimizer-dependent update rule of the form
\begin{equation}
\theta_{t+1} = \theta_t + f\big(\theta_t, \nabla_{\theta_t} \mathcal{L}(X_B, Y_B; \theta_t)\big),
\end{equation}
where $(X_B, Y_B)$ denotes a mini-batch, $\theta_0$ is a random parameter initialization, and $t$ indexes discrete training steps. The Hessian of the loss is given by
$H(\theta) = \nabla^2_{\theta} \mathcal{L}(X,Y;\theta) \in \mathbb{R}^{N \times N}.$
As a symmetric matrix, $H(\theta)$ admits an eigendecomposition with real eigenvalues and orthonormal eigenvectors. Let $\{(\lambda_i, v^{(i)})\}_{i=1}^N$ denote the eigenpairs, ordered such that $\lambda_1 \geq \lambda_2 \geq \dots \geq \lambda_N$.

Note that the Hessian depends on four components: model architecture, parameters, loss function, and dataset. 
To study how curvature evolves during training (with only parameters varying), we track the time-dependent eigenvectors
\begin{equation}
\{ v^{(i)}(t) \}_{i=1}^N \quad \text{from} \quad H(\theta_t),
\end{equation}
denoted equivalently as $v^{(i)}(H({\theta_t}))$, over varying $t$. To quantify the changes we use two metrics: \emph{displacement} and \emph{localization}.

\paragraph{Vector change metric: Displacement} Following \cite{baity-jesi_comparing_2018}, which studies weight dynamics of neural networks compared to glassy systems, we quantify eigenvector change via
\begin{equation}
    \Delta^{(i)} (t_w, t_{w}+t) = \frac{1}{N}||v^{(i)}(t_w) - v^{(i)}(t_w + t)||_2^2.
    \label{eq:displacement}
\end{equation}
This \emph{displacement} is the change of the $i$th eigenvector $v^{(i)}$ between starting time $t_w$ and after waiting time $t$. This acts as a two-time correlation function: it depends only on $t$ in a stationary learning regime, and on both $t$ and $t_w$ otherwise.

As a reference point, we consider the expected Hessian under random parameter initialization
$
\mathbb{E}_{\theta_0 \sim \mathcal{D}}[H(\theta_0)],
$
where $\mathcal{D}$ denotes the initialization distribution. This baseline captures the loss landscape at initialization, and serves as a point of comparison for understanding how optimization reshapes the Hessian eigenspaces. The \emph{displacement baseline} is defined as 
\begin{equation}
    \tau^{(i)} = \displaystyle \mathop{\mathbb{E}}_{
    \theta, \psi \sim \mathcal D
    }\Big[\frac{1}{N} || v^{(i)}(H(\theta)) - v^{(i)}(H(\psi)) ||_2^2\Big].
    \label{eq:dispalcement_baseline}
\end{equation}

\paragraph{Vector Localization: Inverse Participation Ratio.}
 Localization refers to whether the mass of a vector is concentrated on a small subset of coordinates or spread uniformly across many.
We quantify eigenvector localization using the Inverse Participation Ratio (IPR).  For a vector $e \in \mathbb{R}^N$ with $||e||_2=1$,
$IPR(e) = \sum_{j=1}^N e_j^4$
which satisfies $\mathrm{IPR}(e) \in [\frac{1}{N},\, 1]$. 
Small values of IPR of $\mathcal{O}(1/N)$ correspond to \emph{delocalized} vectors, while large values of $\mathcal{O}(1)$ indicate \emph{localization} on a few dominant entries. More generally, $IPR\sim 1/k$ means that the vector is supported on approximately $k$ entries.
IPR is widely used in disordered systems \cite{murphy2011generalized} (e.g., Anderson localization) and random matrix theory \cite{Rudelson_2015, OROURKE2016361}. In machine learning, where hessian eigenvector entries correspond to model parameters, the IPR measures how sensitivity is distributed across the network.

\section{Experiments}
We consider classification on MNIST \cite{mnist} and FashionMNIST \cite{xiao2017fashionmnistnovelimagedataset}. We train a 3-layer fully-connected MLP (widths 100, 100, 10, ReLU) with cross-entropy loss, following \cite{baity-jesi_comparing_2018}. Parameters are Xavier-initialized \cite{glorot_understanding_2010}. Training uses mini-batches of size 128 for $10^5$ iterations. We compare SGD, Adam \cite{kingma2017adammethodstochasticoptimization}, and SAM \cite{foret_sharpness-aware_2021} (with SGD base), using fixed learning rates ($10^{-2}$ for SGD/SAM, $10^{-3}$ for Adam) and standard hyperparameters. Because Hessian dynamics are known to depend on optimization hyperparameters such as batch size and learning rate \cite{yao2018hessian,jastrzebski2019relationsharpestdirectionsdnn,WANG2026108874}, we additionally perform dedicated ablation studies over both quantities in Appendix \ref{app:ablation}. Each run is repeated over 5 random initializations. Standard performance metrics are reported in Appendix \ref{app:training}, as our focus is on training dynamics.

To track loss landscape geometry, we compute the top three Hessian eigenvectors every $\sim$100 steps, yielding 1584 snapshots. The Hessian is evaluated on the full dataset. 
Eigenvectors are computed via Lanczos iteration \cite{lanczos1950iteration} using Hessian-vector products (Pearlmutter’s trick \cite{Pearlmutter_HVP_trick}), without forming the full Hessian.

\subsection{Eigenvector displacement across training}
\label{sec:displacement}

\begin{figure}
    \centering
    \includegraphics[width=1\linewidth]{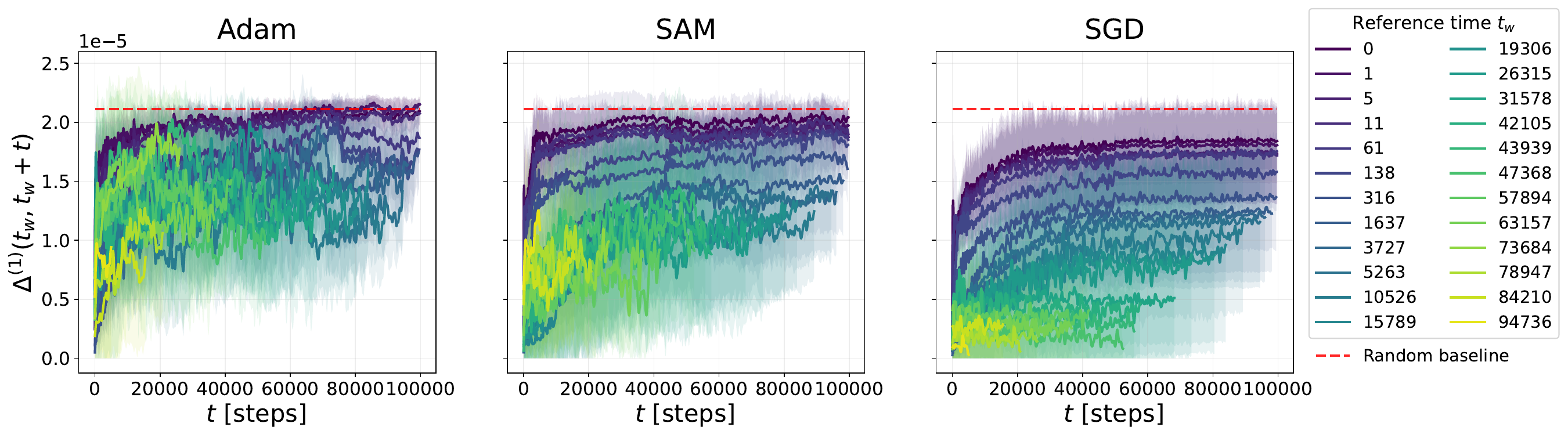}
    \caption{Two-time mean square displacement, $\Delta(t_w, t_w+t)$ as defined in Equation \ref{eq:displacement}. Each curve $t_w$ corresponds to the leading eigenvector at time $t_w$ and its change after waiting $t$ steps. The three plots correspond to the three optimizers. Red dashed line indicates the random baseline.}
    \label{fig:displacement}
\end{figure}

We analyze the displacement of the leading eigenvector $v^{(1)}$ during optimization. Following the two-time correlation analysis of \cite{baity-jesi_comparing_2018}, we use $\Delta(t_w,t_w+t)$ to distinguish stationary (primarily $t$-dependent) from aging dynamics (explicitly dependent on both $t$ and $t_w$). 
To contextualize these values, we compare them to the empirically estimated \emph{displacement baseline} defined in Equation \ref{eq:dispalcement_baseline}, which corresponds to complete decorrelation between eigenvectors.
As shown in Figure ~\ref{fig:displacement}, we observe an \emph{early-time collapse} for small $t_w$ where curves for different waiting times $t_w$ approximately collapse. The $v^{(1)}$ direction changes rapidly, but with little dependence on training age. 
This is followed by a \emph{aging regime} for later $t_w$, where a clear dependence on $t_w$ emerges: change occurs over progressively longer time scales for larger $t_w$, consistent with aging behaviour observed in glassy systems. 
The \emph{late-time behaviour}, for large $t$, the curves flatten. For SGD, saturation occurs below the random baseline, indicating partial stabilization. For Adam, flattening occurs close to the baseline, suggesting near-complete decorrelation. SAM exhibits intermediate behaviour, with slower but persistent drift. These observations extend to training on the FashionMNIST dataset as well (see Figure \ref{fig:app:displacmeent_fashion} in the Appendix).

These observations indicate that optimizers can produce different dynamics in the leading curvature directions. SGD showed a gradual stabilization of $v^{(1)}$, while Adam resulted in persistent shifting throughout training. The differences are robust to variations in both batch size and learning rate, with only the full-batch regime displaying a distinct behaviour (Appendix~\ref{app:ablation}).

\subsection{Alignment with gradient and parameter update}
\label{sec:alignment}

\begin{figure}
\centering
    \centering
    \includegraphics[width=0.8\linewidth]{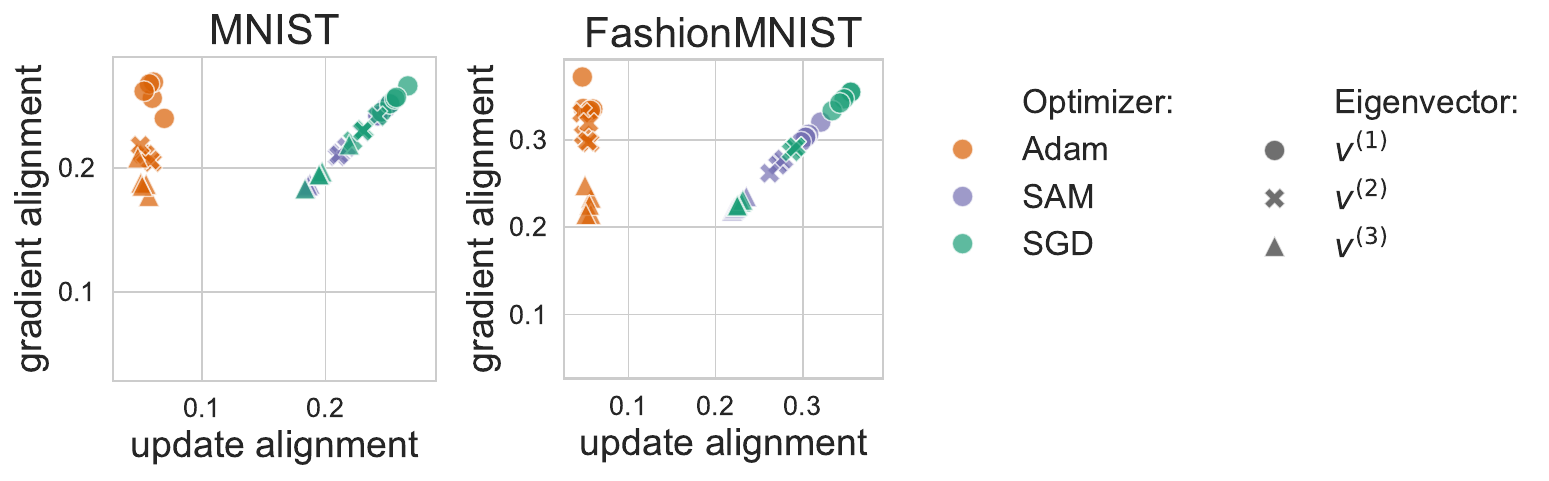}

    \caption{Dot product of normalized gradient and update with $v^{(i)}$ for $i=1,2,3$. Points in the plot correspond to mean alignment values over different experiment initializations. Alignment scales with eigenvalue magnitude. SAM exhibits slightly lower alignment than SGD, with both lying on the diagonal (as expected, since update $\approx$ gradient). In contrast, Adam’s update alignment remains under $0.1$, likely related to the momentum parameter $\beta_1=0.9$ used during training.}
    \label{fig:alignment}
\end{figure}
One possible explanation of the above optimizer difference relates to how each optimizer uses gradient information. SGD follows the gradient, SAM incorporates local sharpness to the gradient-based update, and Adam uses adaptive, momentum-like updates. Prior work \cite{gur-ari_gradient_2018} has shown that gradients largely lies in the span of the top Hessian eigenvectors, and that strong alignment with leading eigenvectors can stall optimization \cite{song_does_2025}. Alignment with dominant eigenvectors appears to reduce optimization speed \emph{and to stabilize} these directions. This is consistent with a velocity interpretation of second-order derivatives: moving along strongly convex directions reduces gradient magnitude (and thus slows updates), while movement along flatter or concave directions does not.
Because the gradient/update alignments tend to have a stable trend throughout training, we summarize them by reporting the mean over training time - see Figure \ref{fig:alignment} with raw values reported in Appendix \ref{app:alignment}.

\subsection{Eigenvector Localization - Inverse Participation Ratio}
\label{sec:IPR}

\begin{figure}[h]
\centering
\begin{minipage}{0.45\textwidth}
    \centering
     \includegraphics[width=\linewidth]{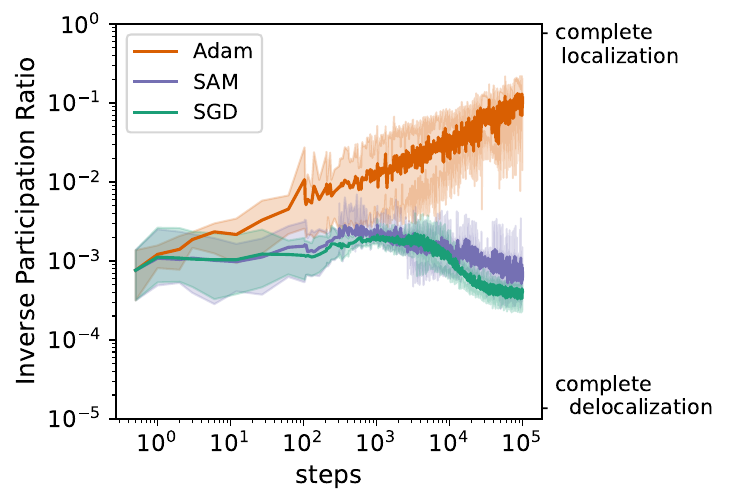}
    \caption*{(a) MNIST dataset}
\end{minipage}
\hfill
\begin{minipage}{0.45\textwidth}
    \centering
    \includegraphics[width=\linewidth]{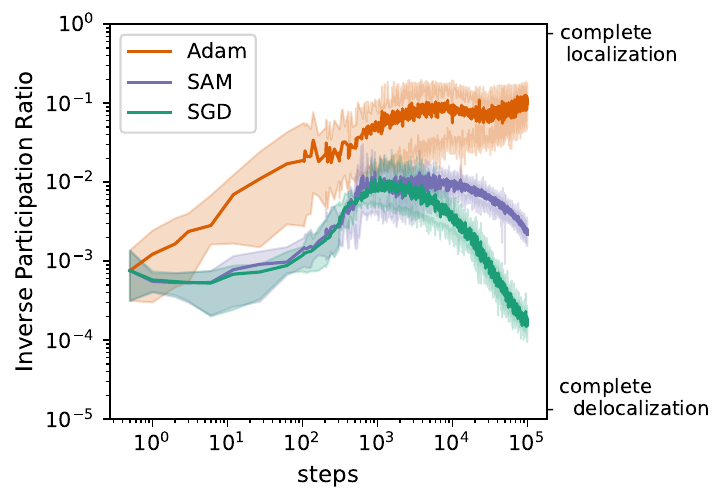}
    \caption*{(b) FashionMNIST dataset}
\end{minipage}
\caption{Inverse Participation Ratio of the leading eigenvector throughout training.  Mean and min-max over 5 initializations. The training using Adam progressively localizes the eigenvectors while SGD-based optimizers have a delocalizing trend with a very small variation between runs.}
\label{fig:ipr}
\end{figure}
We track the leading eigenvector’s localization during training (Figure \ref{fig:ipr}). Clear differences emerge across optimizers: under Adam, the eigenvector becomes strongly localized (IPR $\approx 0.1$), whereas SGD progressively delocalizes toward the uniform baseline ($1/N = 10^{-5}$) with low variance across runs. SAM exhibits slightly higher localization than SGD while maintaining similarly low variance.  These trends are robust across most learning-rate and batch-size settings, and are most pronounced for large learning rates and small-to-moderate batch sizes, while breaking down in the full-batch regime (see Appendix \ref{app:ablation}). Overall, these distinct localization behaviours indicate that different optimizers explore qualitatively different regions of the loss landscape.

In the previous section, we observed substantial diffusion of eigenvectors under the Adam optimizer. Yet despite this ongoing mixing, the leading eigenvector localization remains stable along a consistent trend.
This suggests that, even as the top eigenvectors rotate, they share a localization value throughout training (see Figure \ref{fig:app:15eig_IPR} in the Appendix for the top fifteen eigenvectors).
Since the Hessian encodes parameter sensitivity, this localization may carry practical significance for targeted pruning and uncertainty quantification \cite{wang_eigendamage_2019}. Recall that eigenvector components correspond directly to model parameters; thus, a localized eigenvector indicates that a small subset of weights dominates the most convex direction (see Figure \ref{fig:app:reshaped} in the Appendix for a visualization).
    
\section{Conclusion}

In this work, we studied the dynamics of leading Hessian eigenvectors during neural network training, focusing on their displacement over time and localization across parameters. In the displacement dynamics, we observe aging-like behaviour and optimizer-dependent stabilization, with SGD leading to curvature stabilization and Adam to continued eigenvector mixing. 
Localization metric further refines this picture, showing that different optimizers explore qualitatively distinct regions of the loss landscape. Moreover, localization appears to be a stable statistic throughout training: it shows very low variance across runs, remains stable even when eigenvectors undergo significant mixing, and sometimes continues to exhibit a clear trend at convergence rather than saturating. These effects are robust to learning-rate and batch-size variations (Appendix~\ref{app:ablation}), with deviations primarily limited to the full-batch and small-learning-rate regimes.

Future work could leverage these observations to improve optimizer dynamics and exploit parameter redundancy. Eigenvector mixing may relate to edge-of-stability, though mostly studied in full-batch settings, these observations may help extend it to mini-batch regimes.

\section*{Acknowledgements}
This work was supported by the Engineering and Physical Sciences Research Council (EPSRC) [grant number EP/Z534882/1] through the \emph{EIGENDATA} project.

\bibliography{bib}
\newpage
\clearpage
\appendix

\section{Related work}
\paragraph{Gradient alignment with the top Hessian subspace.}

A growing body of work has identified a low-rank structure in the Hessian of deep classification models. Spectral analyses show that during training, a small number of outlier eigenvalues separate from the bulk of the spectrum, with the number of dominant modes often scaling with the number of classes \cite{sagun2017eigenvalueshessiandeeplearning,gur-ari_gradient_2018}.

Building on these observations, \cite{jastrzebski2019relationsharpestdirectionsdnn} showed that SGD updates become strongly aligned with directions of sharp curvature during training. In particular, \cite{gur-ari_gradient_2018} demonstrated that the gradient lies within the subspace spanned by the top Hessian eigenvectors in classification settings.  Within this subspace, however, the gradient does not exhibit a preferred direction and appears approximately isotropic with respect to the eigenvector basis. Recent theoretical work has also established alignment between SGD updates and dominant Hessian subspaces in simplified settings, including multi-class logistic regression and shallow neural networks \cite{arous2024highdimensional}.

Subsequent work connected this phenomenon to broader optimization dynamics. \cite{fort_emergent_2019} related gradient concentration in the top Hessian subspace to the emergence of low-dimensional learning dynamics and edge-of-stability behaviour. Similarly, \cite{ghorbani2019investigationneuralnetoptimization} argued that outlier eigenvalues can induce a form of \emph{gradient concentration}, where optimization becomes dominated by a small number of curvature directions.

Recent work by \cite{song_does_2025} suggests that the sharpest eigendirections are not necessarily the most informative for descent. In particular, they show that restricting optimization to the leading Hessian eigenspace stalls learning. They suggest that useful updates may occur primarily in flatter directions near the Hessian null space.

\paragraph{Stability of the top Hessian eigenspace}

Several works have observed that the leading Hessian eigenspace remains relatively stable throughout training \cite{alain_negative_2019,gur-ari_gradient_2018}. This phenomenon, sometimes referred to as \emph{Hessian inertia}, has been connected to feature learning dynamics \cite{bao_hessian_inertia_2023}.
Due to eigenvector mixing, prior work typically studies the stability of the dominant \emph{subspace} rather than individual eigenvectors. Existing approaches quantify this stability using measures such as greedy cosine similarity, Rayleigh quotient or subspace-overlap metrics \cite{alain_negative_2019, bao_hessian_inertia_2023, allez_eigenvector_2012}.

More recently, a connection has been identified between eigenvector rotations and the discovery of flat solutions \cite{WANG2026108874}. In particular, the authors introduce the \emph{Rotational Polarity of Eigenvectors} (RPE), a quantity that captures geometric rotations of the leading Hessian eigenvectors. Above a stability threshold, increased RPE is associated with more exploratory optimization dynamics. This effect becomes more pronounced at larger learning rates.

\paragraph{Flatness and Sharpness-Aware Minimization (SAM)}

Empirical studies consistently observe that the Hessian spectrum contains many eigenvalues near zero. These directions correspond to flat regions of the loss landscape and may connect different minima \cite{granziol_beyond_2021}. Flat minima, characterized by many near-zero Hessian eigenvalues, have also been associated with improved generalization and robustness to noise \cite{dinh2017sharp, keskar2017largebatchtrainingdeeplearning, jiang2020fantastic}.

Motivated by these observations, \cite{foret_sharpness-aware_2021} introduced Sharpness-Aware Minimization (SAM), an optimization method that implicitly incorporates local curvature information without explicitly computing the Hessian. SAM minimizes the worst-case loss within a small neighbourhood in parameter space, encouraging convergence toward flatter solutions. The method has been highly effective across a broad range of architectures and applications, including transformers and large language models. Subsequent work has further analysed the optimization dynamics and convergence properties of SAM, including connections to the edge-of-stability regime \cite{agarwala_sam_2023}.




\paragraph{Hessian as a proxy for parameter sensitivity}

Since the eigenvectors of the Hessian define principal curvature directions of the loss landscape, they provide a natural local representation of parameter sensitivity. Second-order methods have long exploited this idea, using curvature to estimate the effect of parameter perturbations. In particular, the leading eigenvectors correspond to directions along which the loss changes most rapidly, while near-zero eigenvalues indicate flat directions that are more tolerant to perturbations and may admit parameter sparsification \cite{wang_eigendamage_2019, dong2019hawq}. From a Bayesian perspective, the Hessian of the log posterior defines the covariance of a local Gaussian around the MAP estimate, linking flatter directions in the loss landscape to higher posterior uncertainty \cite{husmeier2000bayesian}.

\paragraph{Inverse Participation Ratio}
Primarily used in random matrix theory and statistical physics to quantify eigenvector localization, the inverse participation ratio (IPR) has only recently begun to appear in machine learning contexts. To our knowledge, its application to Hessian eigenvectors remains largely unexplored. Recent work \cite{NGUYEN2026131474} instead applies the IPR to weight matrix eigenvectors, showing that it acts as a diagnostic of learning regimes and reveals a phase transition linking localization strength to task complexity. 

\section{Experimental details}
\label{app:training}

We provide additional details on the experimental setup, evaluation metrics, and computational procedures used in the main paper. This appendix section is organized as follows: Section~\ref{app:setup} summarizes the experimental configuration, Section~\ref{app:metrics} reports the evaluation metrics, Section~\ref{app:methods} describes the computation of Hessian-based quantities, and Section~\ref{app:extras} provides implementation choices and additional observations.

\subsection{Experimental setup}
\label{app:setup}

We summarize the experimental configuration used throughout the paper:

\begin{itemize}
    \item \textbf{Datasets:} MNIST \cite{mnist} and FashionMNIST \cite{xiao2017fashionmnistnovelimagedataset} for multi-class classification.

    \item \textbf{Model:} A 3-layer fully-connected MLP with layer widths $(100, 100, 10)$ and ReLU activations, following \cite{baity-jesi_comparing_2018}. Parameters are initialized using Xavier initialization \cite{glorot_understanding_2010}.

    \item \textbf{Training setup:} Cross-entropy loss optimized using mini-batch training with batch size 128 for $10^5$ iterations.

    \item \textbf{Optimizers:} We compare SGD, Adam \cite{kingma2017adammethodstochasticoptimization}, and SAM \cite{foret_sharpness-aware_2021} (with SGD as base optimizer).

    \item \textbf{Learning rates:} Fixed throughout training: $10^{-2}$ for SGD and SAM, and $10^{-3}$ for Adam. Standard optimizer hyperparameters are used in all cases.

    \item \textbf{Repetitions:} Each experiment is repeated over 5 random initializations to account for stochasticity.

    \item \textbf{Hessian eigenvectors:} We compute the top-3 Hessian eigenvalues and eigenvectors every $\sim$100 training steps. For a selected seed, we additionally compute the top-15 eigenvalues to obtain a finer spectral resolution.

\end{itemize}

\subsection{Metrics and evaluation protocol}
\label{app:metrics}

We now report the optimization performance and Hessian-based measures describing loss landscape geometry.

\begin{figure}[h]
\centering

\begin{minipage}{0.49\textwidth}
    \centering
    \includegraphics[width=\linewidth]{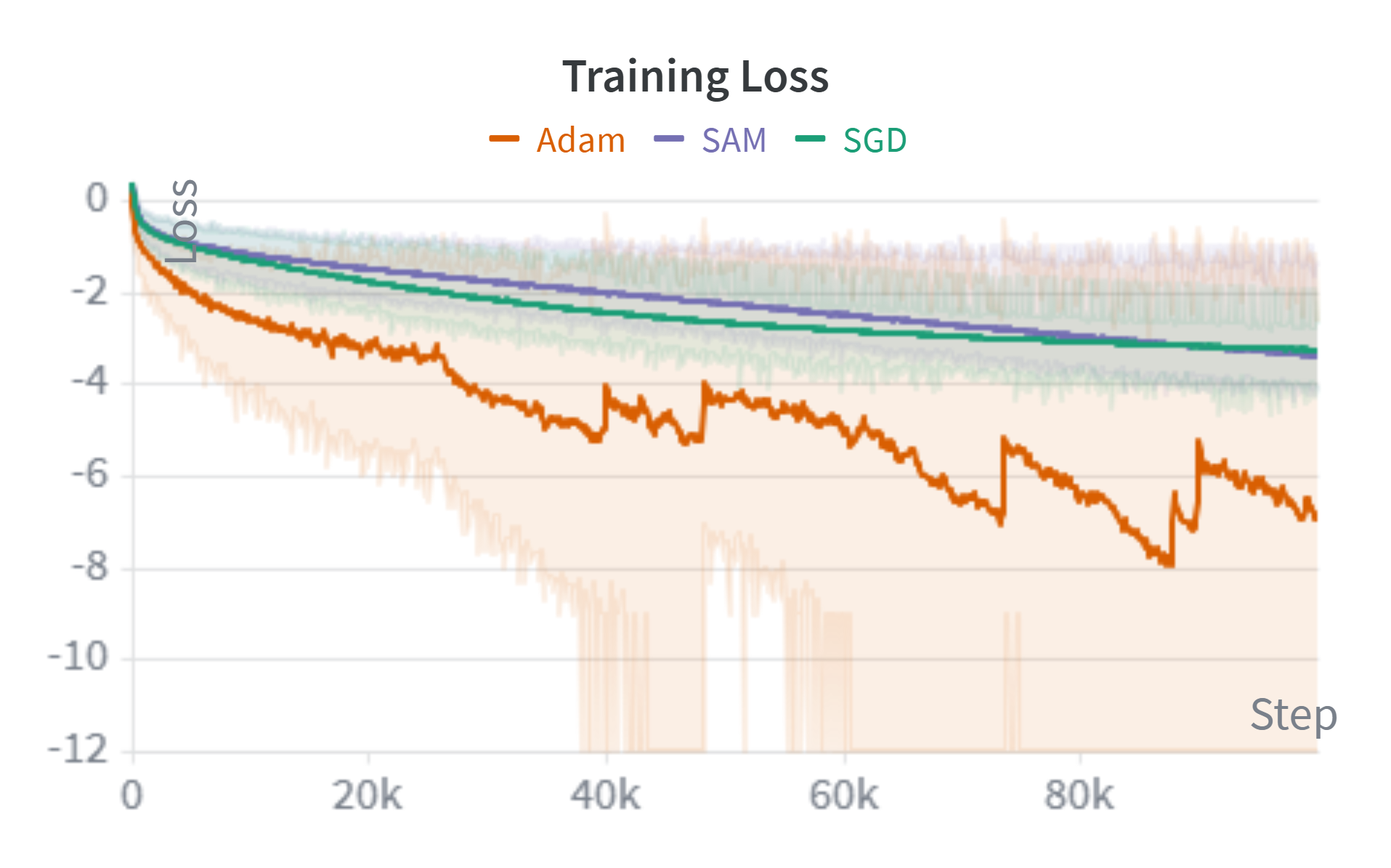}
    \caption*{MNIST dataset}
\end{minipage}
\hfill
\begin{minipage}{0.49\textwidth}
    \centering
    \includegraphics[width=\linewidth]{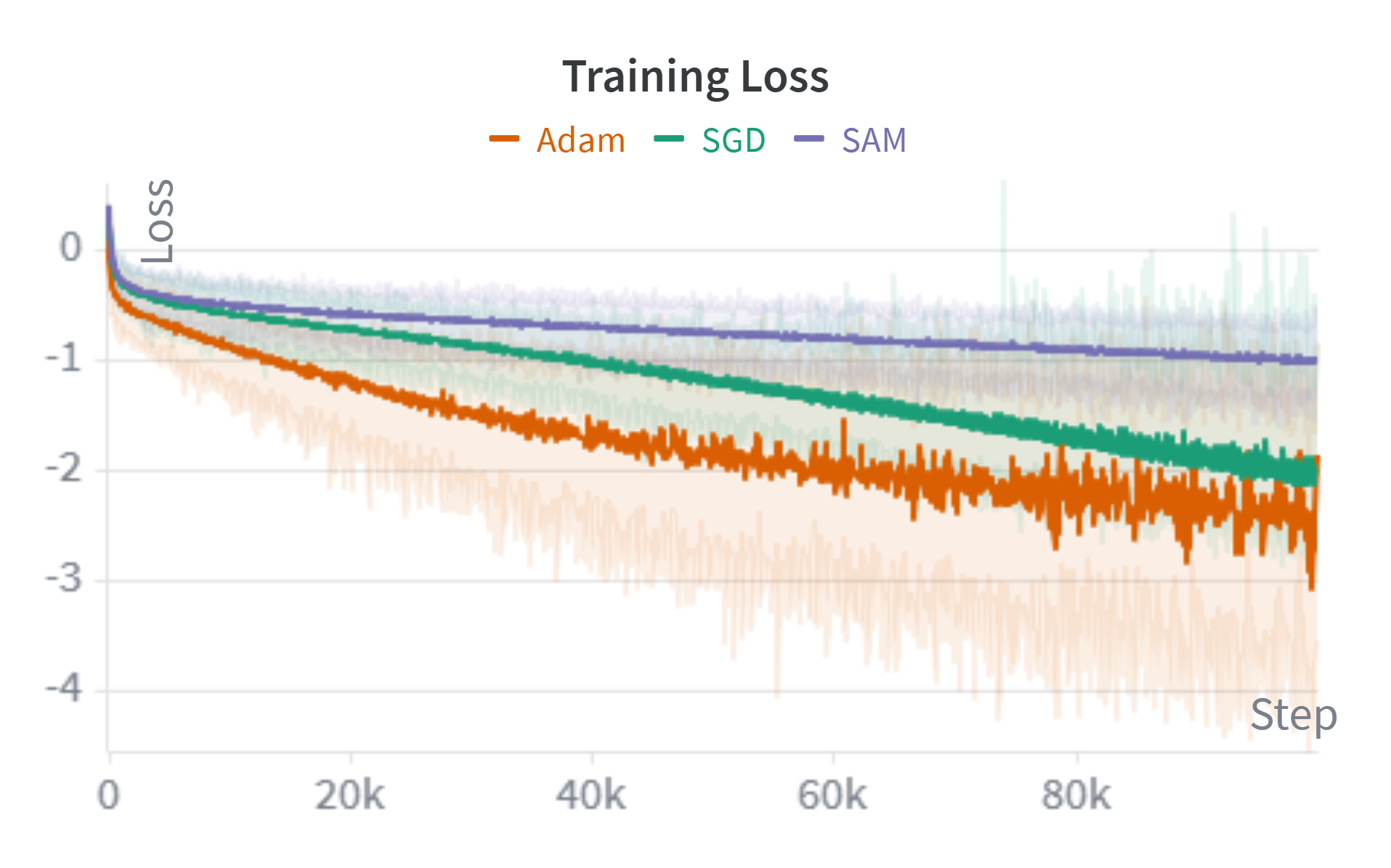}
    \caption*{FashionMNIST dataset}
\end{minipage}
   \caption{Training loss (log-scale). Mean and min-max over 5 random initializations. Adam achieves lowest loss values with notable jumps in MNIST dataset training.}
    \label{fig:app:loss}
\end{figure}

\paragraph{Training Loss} Figure \ref{fig:app:loss} shows the cross-entropy training loss function throughout training for 5 random seeds for all optimizers. In addition to training loss, we report test accuracy to monitor generalization throughout training.

\begin{figure}[h]
\centering
\begin{minipage}{0.4\textwidth}
    \centering
    \includegraphics[width=\linewidth]{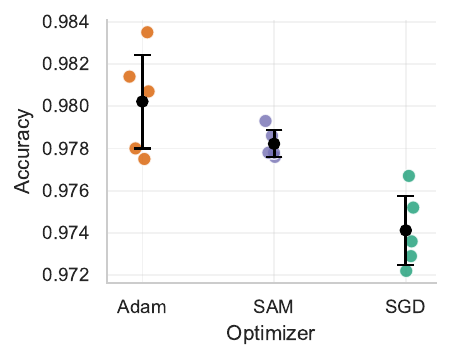}
    \caption*{MNIST dataset}
\end{minipage}
\hfill
\begin{minipage}{0.4\textwidth}
    \centering
    \includegraphics[width=\linewidth]{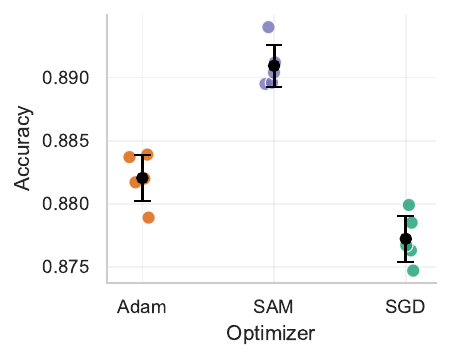}
    \caption*{FashionMNIST dataset}
\end{minipage}
   \caption{Test accuracy across multiple runs, where each point in the scatter plot corresponds to a different random seed initialization. SGD consistently achieves the lowest performance in both cases. Adam attains the highest accuracy on MNIST, while SAM performs best by a clear margin on FashionMNIST.}
    \label{fig:app:accuracy}
\end{figure}

\paragraph{Test accuracy} Figure \ref{fig:app:accuracy} presents the test accuracies grouped by optimizer. Note that since we used $10^5$ training steps without early stopping, this might have led to overfitting. To characterize the geometry of the loss landscape, we report spectral quantities of the Hessian during training.

\begin{figure}[h]
\centering
\begin{minipage}{0.9\textwidth}
    \centering
    \includegraphics[width=\linewidth]{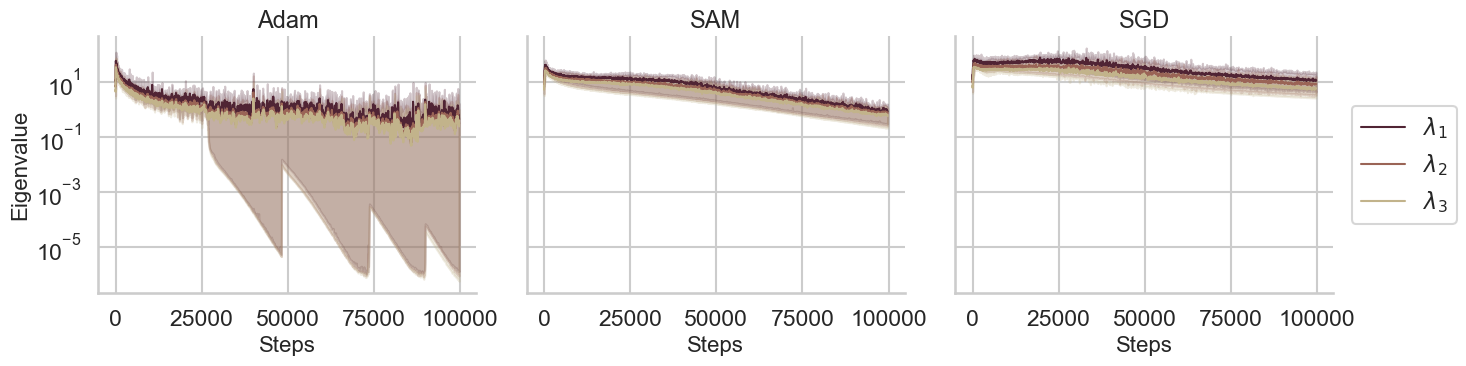}
    \caption*{(a) MNIST dataset}
\end{minipage}
\hfill
\begin{minipage}{0.9\textwidth}
    \centering
    \includegraphics[width=\linewidth]{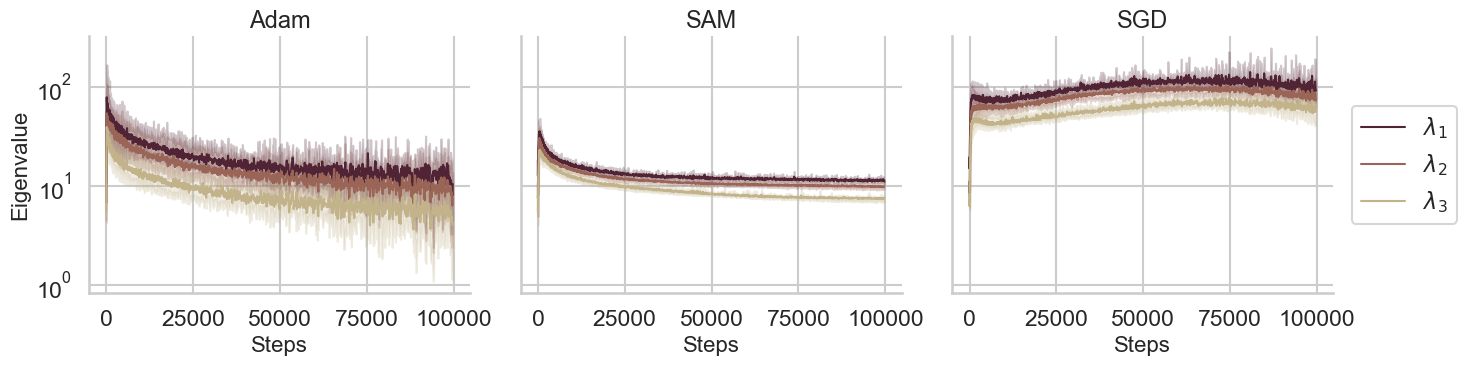}
    \caption*{(b) FashionMNIST dataset}
\end{minipage}
   \caption{Top-3 hessian eigenvalues throughout training. Mean and min-max shown over random initializations.}
    \label{fig:app:spectrum}
\end{figure}

\paragraph{Top Hessian Spectrum} We also plot the recorded top-3 hessian eigenvalues. As seen in Figure \ref{fig:app:spectrum}, eigenvalues for FashionMNIST are larger, reflecting the more complex landscape.

\begin{figure}[h]
\centering
\begin{minipage}{0.3\textwidth}
    \centering
    \includegraphics[width=\linewidth]{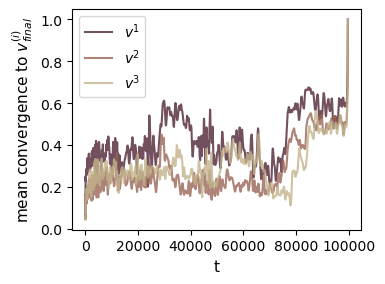}
    \caption*{MNIST - Adam}
\end{minipage}
\begin{minipage}{0.3\textwidth}
    \centering
    \includegraphics[width=\linewidth]{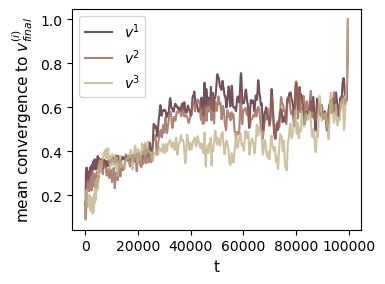}
    \caption*{MNIST - SAM}
\end{minipage}
\begin{minipage}{0.3\textwidth}
    \centering
    \includegraphics[width=\linewidth]{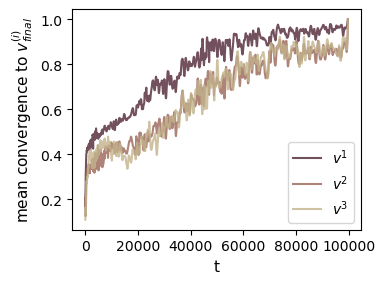}
    \caption*{MNIST - SGD}
\end{minipage}
\hfill
\begin{minipage}{0.3\textwidth}
    \centering
    \includegraphics[width=\linewidth]{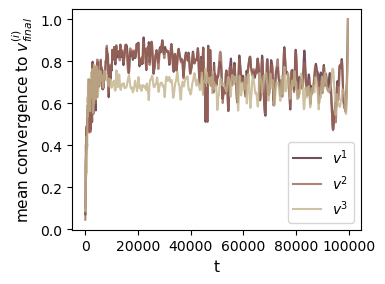}
    \caption*{FashionMNIST - Adam}
\end{minipage}
\begin{minipage}{0.3\textwidth}
    \centering
    \includegraphics[width=\linewidth]{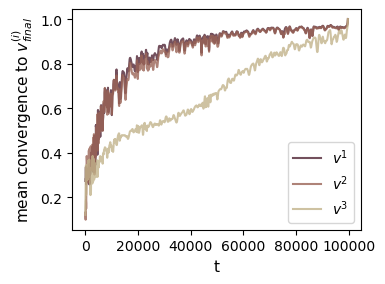}
    \caption*{FashionMNIST - SAM}
\end{minipage}
\begin{minipage}{0.3\textwidth}
    \centering
    \includegraphics[width=\linewidth]{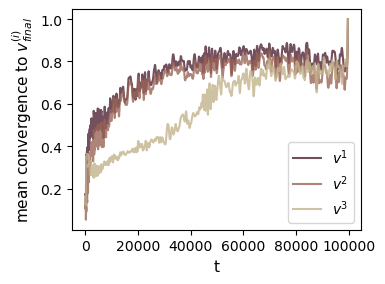}
    \caption*{FashionMNIST - SGD}
\end{minipage}
\hfill

\caption{Convergence of the top-3 eigenvectors to the final minimum basin. This is dot product of $v^{(i)}(t)$ with the final training value $v^{(i)}(t_{\text{final}})$. Average is taken over 5 runs and Gaussian smoothing with $\sigma=2$ applied.}
\label{fig:app:convergence}
\end{figure}

\paragraph{Convergence to final eigenvector}
Assuming the algorithms converge to a minimum, we consider the final set of eigenvectors $v^{(i)}(t_{\text{final}})$ that characterize the basin of this solution. We also analyze how smoothly each eigenvector converges to its final orientation, as well as whether lower-index eigenvectors converge faster, by tracking the cosine similarity $v^{(i)}(t_{\text{final}})^T v^{(i)}(t)$ over $t \in [0, t_{\text{final}}]$. Figure \ref{fig:app:convergence} shows the resulting convergence patterns for different optimizers and the two datasets considered. In many cases, a similarity of $0.8$ is reached within the first few steps. SGD exhibits relatively smooth convergence, although in several runs the similarity abruptly jumps to $1.0$ only at the final iterations rather than increasing gradually, likely due to continual eigenvector mixing. These results complement the displacement plots in Figures \ref{fig:displacement} and \ref{fig:app:displacmeent_fashion}. In most cases, we also observe that $v^{(1)}$ stabilizes earlier than $v^{(3)}$.

\begin{figure}[h]
\centering
\begin{minipage}{0.8\textwidth}
    \centering
    \includegraphics[width=\linewidth]{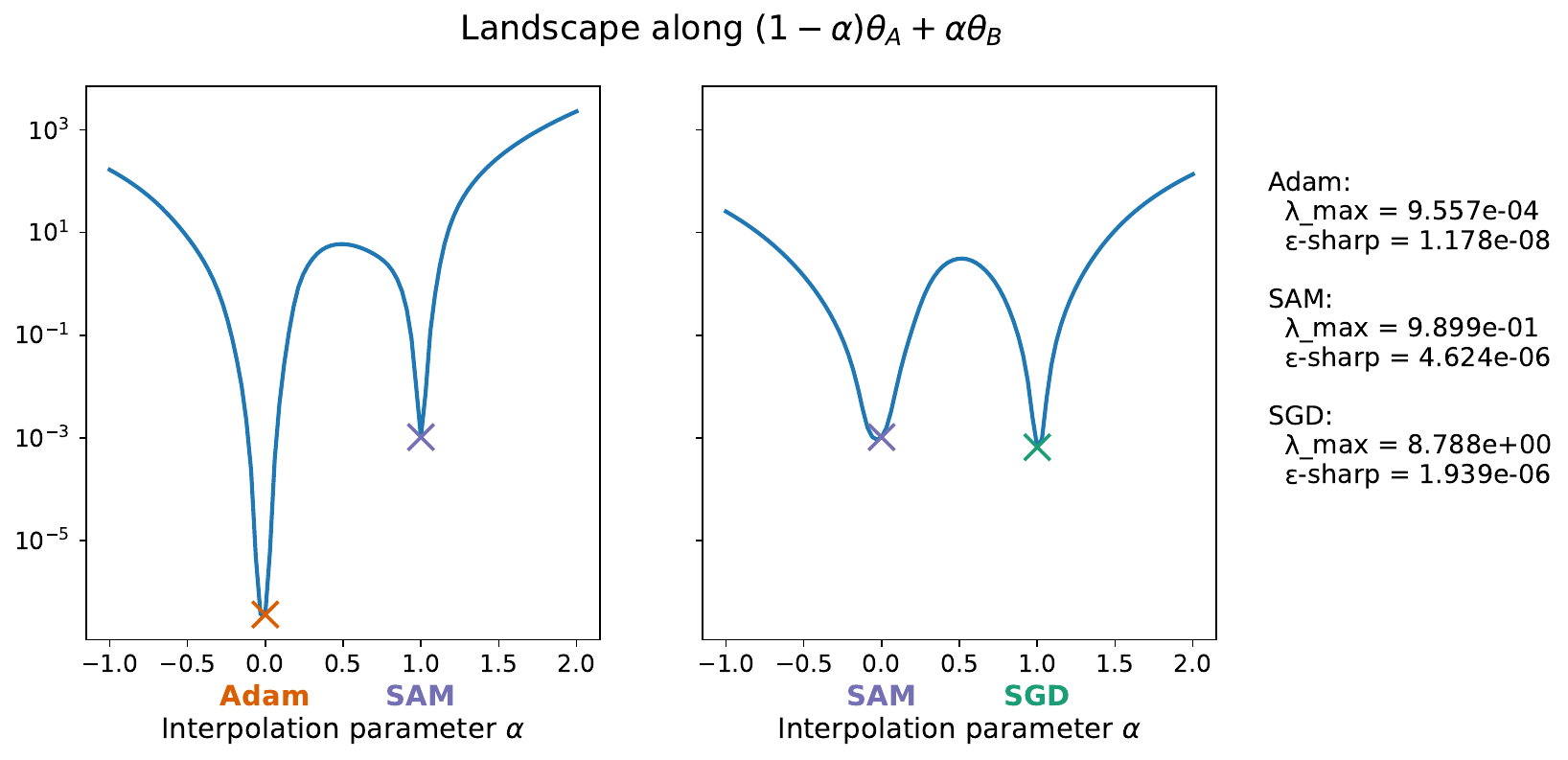}
    \caption*{(a) MNIST solutions}
\end{minipage}
\begin{minipage}{0.8\textwidth}
    \centering
    \includegraphics[width=\linewidth]{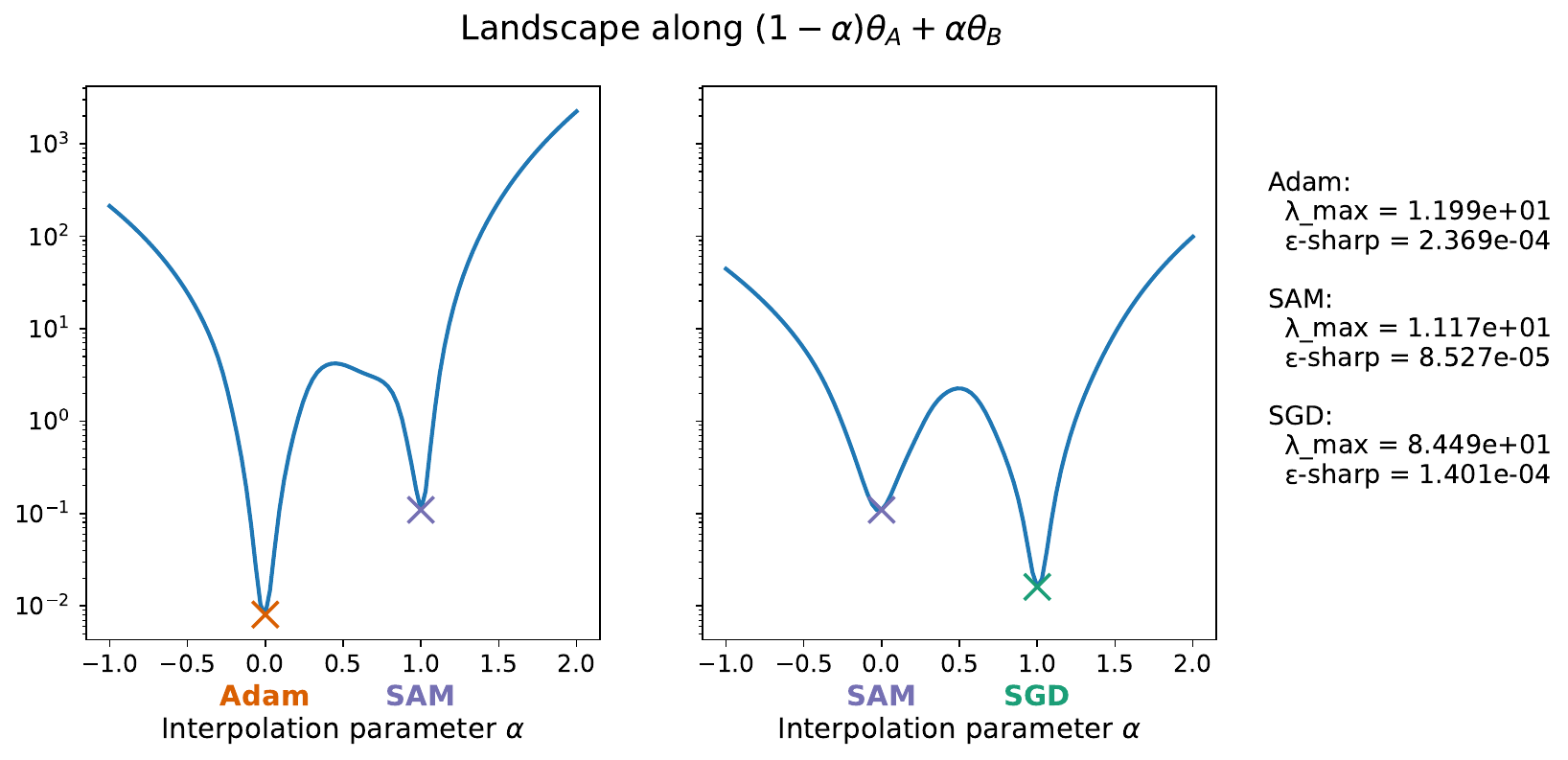}
    \caption*{(b) FashionMNIST solutions}
\end{minipage}

\caption{Comparison of best minima (over 5 runs) between optimizers. We consider the test+train loss on the line interpolating between two different solutions, as defined in Equation \ref{eq:app:minima_connection}. Sharpness metrics such as largest eigenvalue and the $\epsilon$-sharpness are reported on the right-hand side.}
\label{fig:app:minima}
\end{figure}

\paragraph{Connection of the minima}
To assess whether different optimizers converge to qualitatively distinct minima, we visualize the loss landscape between solutions. From five repeated runs per optimizer, we select the checkpoint with the highest test accuracy. We then evaluate the loss (computed over the full dataset) along linear interpolations between two minima $\theta_A$ and $\theta_B$:
\begin{equation}
\alpha\theta_A + (1-\alpha)\theta_B \quad \text{for } \alpha \in [-1, 2].
\label{eq:app:minima_connection}
\end{equation}

This provides insight into the geometry of the loss landscape, in particular the flatness of the minima and the height of the barriers separating them, as shown in Figure \ref{fig:app:minima}. We additionally report the largest Hessian eigenvalue and $\epsilon$-sharpness, which are commonly used proxies for local curvature and flatness \cite{keskar2017largebatchtrainingdeeplearning, dinh2017sharp}. We observe that a lower training loss does not necessarily correspond to higher test accuracy (see Figure \ref{fig:app:accuracy}). In contrast, sharpness appears to be more strongly correlated with generalization performance.

\subsection{Computation of the Hessian eigenvectors}
\label{app:methods}
 Direct computation of the Hessian is expensive and sometimes  infeasible. However, automatic differentiation enables efficient Hessian-vector products using the classical result of \cite{Pearlmutter_HVP_trick}, allowing second-order information to be accessed without explicitly forming the matrix.

We use the \texttt{eigsh} function from the \texttt{scipy.linalg} library, avoiding explicit construction of the Hessian by instead providing a function that computes Hessian–vector products. This amounts to performing a Lanczos iteration to approximate the leading eigenvalues and eigenvectors \cite{lanczos1950iteration}. The specified tolerance of $10^{-3}$ refers to the accuracy of the eigenvalue estimates. Consequently, when eigenvalues are closer than this threshold, the corresponding eigenvectors may also become unreliable or less accurately resolved. For the Adam MNIST experiment, as well as for the computation of the top-15 eigenvectors, machine precision tolerance was used instead.

Recall that eigenvectors are not unique: if $v$ is an eigenvector, then $-v$ is also a valid eigenvector. Moreover, when two or more eigenvalues coincide or are very close, the associated eigenspace is not one-dimensional, and any orthonormal basis within it is valid. In our setting, we focus on the top eigenvectors, whose eigenvalues are typically well separated, so handling the sign ambiguity is sufficient. Accordingly, our similarity metrics are designed to be sign-invariant. In particular, although the displacement in Eq.~\eqref{eq:displacement} is defined using the squared $\ell_2$ distance, in practice we compute
\[
\Delta(v,w)
= \frac{1}{N}\min\left\{
\|v - w\|_2^2,\,
\|v + w\|_2^2
\right\},
\]
thereby selecting the closer alignment under the symmetry $v \sim -v$.
We also note that eigenvector mixing (i.e., changes in ordering or rotations during training) is natural and expected in this regime.

\subsection{Justification of design choices}
\label{app:extras}

We conclude with additional implementation decisions and robustness considerations.

\begin{itemize}
    \item Keeping the learning rate fixed during training. We are aware that this is suboptimal and may be too large in later stages of training but we decide not to use early stopping or learning rate scheduler to be able to observe the different stages of training including diffusion around a minimum at the end of training.
    \item Using ReLU activation. It is known that ReLU can stop gradient propagation and so influence the structure of Hessian (that is calculated using 2 gradient passes). Many works use the Sinh activation instead to get smoother spectrum estimates. In our case, however, we observe stable and interpretable results with ReLU. Given its stronger empirical performance and popularity, we retain it for practical relevance.
    \item MLP with 3 fully connected layers 100, 100, 10 was used in \cite{baity-jesi_comparing_2018}. While the metrics considered in this work generalize to more complex architectures, we first focus on this simpler setting as a controlled benchmark for understanding the underlying phenomena.
    \item The digit-recognition MNIST dataset was used in \cite{baity-jesi_comparing_2018}. We extend the validation of these results by considering its more challenging counterpart, Fashion-MNIST \cite{xiao2017fashionmnistnovelimagedataset}. Both datasets share the same image dimensions, number of classes, and number of samples, with the primary difference lying in task complexity. This consistency allows us to maintain eigenvectors of identical size and enabling more meaningful interpretations.
    \item Tracking single eigenvectors. A natural criticism of tracking individual eigenvectors is that they may mix rapidly during training. While we agree that subspace-level analysis is clearer, individual eigenvectors still offer insight into \emph{whether} mixing occurs and \emph{how quickly} it does. Indeed, recent work \cite{WANG2026108874} has liked eigenvector mixing with increased exploration and flatter minima.
    \item Eigenvectors are calculated with precision $1e-3$ since the gaps between top 3 eigenvalues are sufficiently big (see Figure \ref{fig:app:spectrum}). We increase the accuracy for MNIST - Adam combination because we observe the gap dropping below $1e-3$. We also use the machine-precision accuracy when computing the top 15 eigenvalues/-vectors.

\end{itemize}

\clearpage

\section{Displacement - additional experiments}
\label{app:displacement}
We provide additional plots complementing Section \ref{sec:displacement}.

\begin{figure}[h]
    \centering
    \includegraphics[width=1\linewidth]{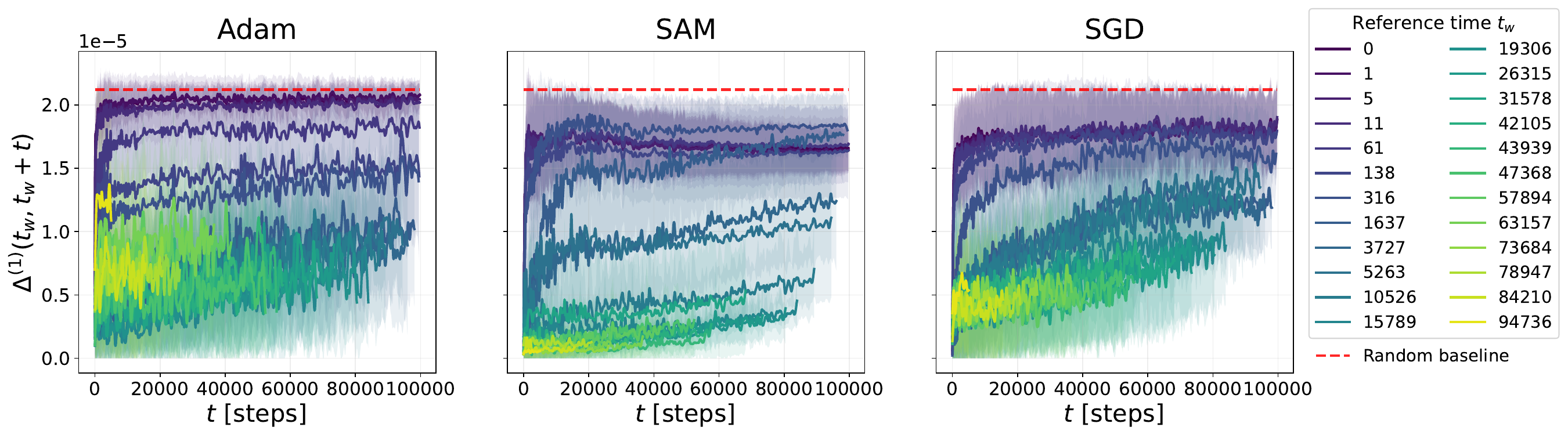}
    \caption{Training on the FashionMNIST dataset. Two-time mean square displacement, $\Delta^{(i)}(t_w, t_w + t)$ as defined in Equation \ref{eq:displacement}. Each curve
$t_w$ corresponds to the leading eigenvector at time $t_w$ and its diffusion after waiting $t$ steps. The three
plots correspond to the three optimizers. Red dashed line indicates the random baseline.}
    \label{fig:app:displacmeent_fashion}
\end{figure}

Figure \ref{fig:app:displacmeent_fashion} shows the displacement $\Delta^{(1)}(t_w, t_w + t)$ during training on the FashionMNIST dataset. Note that the random baseline $\tau^{(1)}$, indicated by the red line, has been re-computed to account for data distribution influence on the Hessian baseline model. Many of the observations from the main text still hold in this more challenging setting, in particular Adam reaching the random baseline while SGD does not, as well as stronger diffusion under Adam.

\begin{figure}[h]
\centering
\begin{minipage}{0.9\textwidth}
    \centering
    \includegraphics[width=\linewidth]{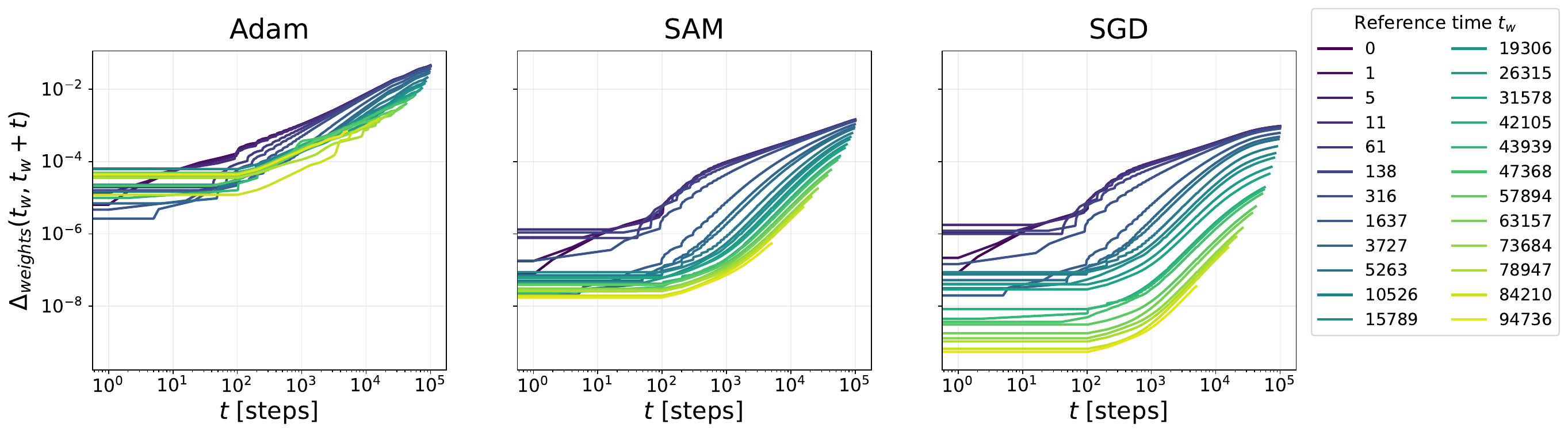}
    \caption*{(a) MNIST}
\end{minipage}
\hfill
\begin{minipage}{0.9\textwidth}
    \centering
    \includegraphics[width=\linewidth]{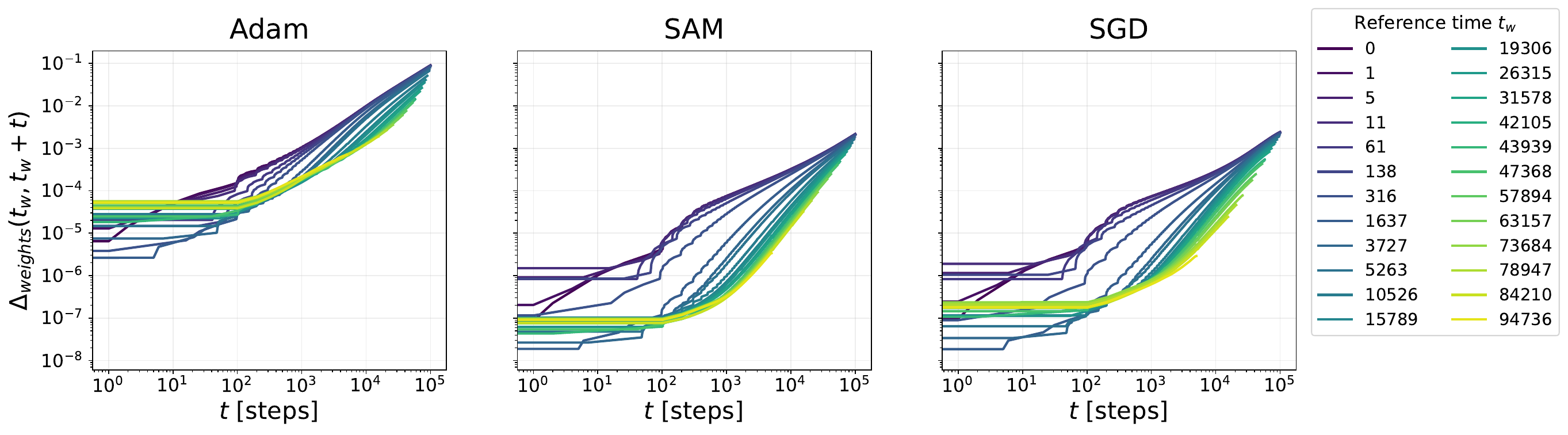}
    \caption*{(b) FashionMNIST}
\end{minipage}

   \caption{Two-time weight displacement metric $\Delta_{weights}(t_w, t_w+t)$ as defined in \cite{baity-jesi_comparing_2018}, without normalization by gradient noise.}
    \label{fig:app:weight_displacement}
\end{figure}

Figure \ref{fig:app:weight_displacement} presents the weight displacement as defined in the original work \cite{baity-jesi_comparing_2018}. Unlike the original formulation, we do not normalize the displacement by gradient noise, which likely explains why we do not observe the flattening at large $t$. We observe a strong similarity between MNIST and FashionMNIST for the same optimizer, with most differences arising at later waiting times $t_w$.

\begin{figure}[h]
\centering
\begin{minipage}{0.9\textwidth}
    \centering
    \includegraphics[width=\linewidth]{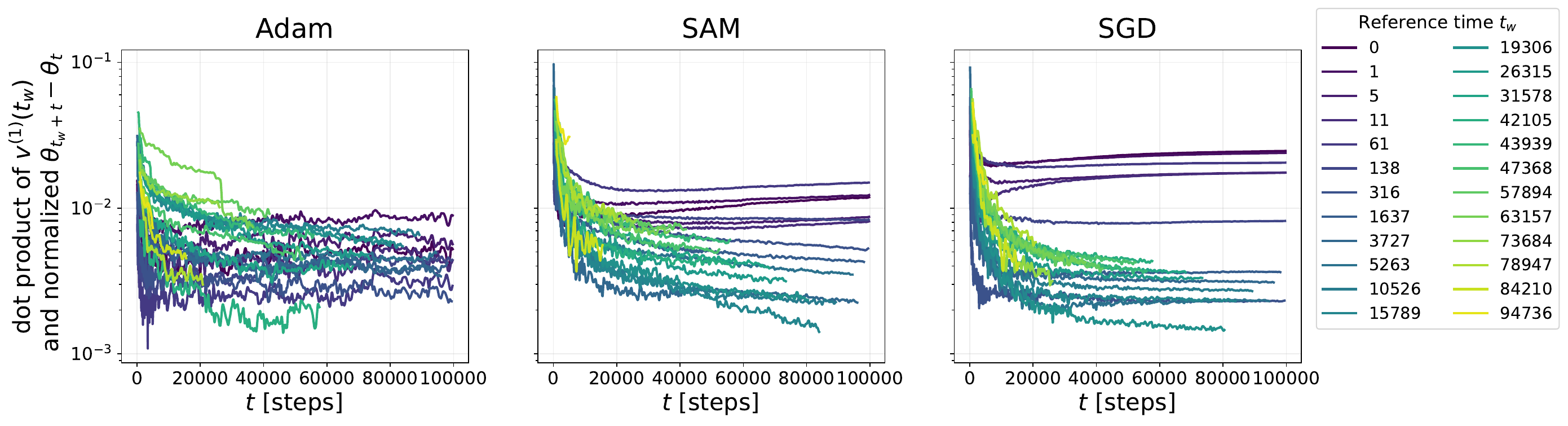}
    \caption*{(a) MNIST}
\end{minipage}
\hfill
\begin{minipage}{0.9\textwidth}
    \centering
    \includegraphics[width=\linewidth]{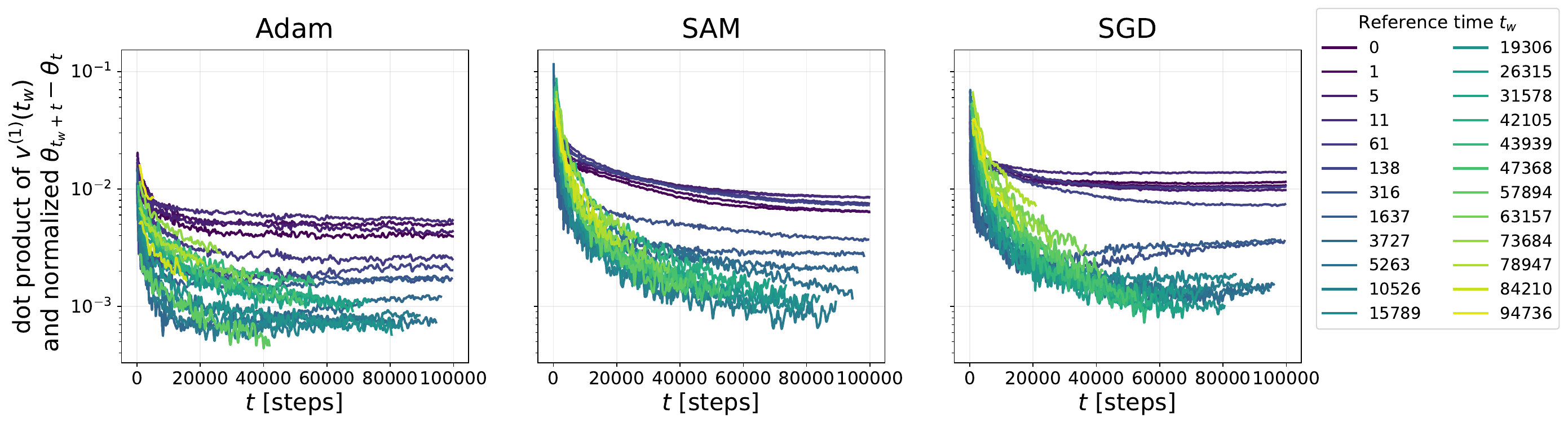}
    \caption*{(b) FashionMNIST}
\end{minipage}

   \caption{The alignment of weight change with the leading eigenvector direction. More precisely, it is a dot product of $v^{(i)}(t_w)$ with the normalized vector of weight change $\theta_{t_w+t}-\theta_t$.}
    \label{fig:app:alignment_weight_change}
\end{figure}

Figure \ref{fig:app:alignment_weight_change} presents the alignment of the weight change direction and leading eigenvector. More precisely,
\begin{equation}
\Big|v^{(1)}(t_w)^T \frac{\theta_{t_w+t} - \theta_{t_w}}{||\theta_{t_w+t} - \theta_{t_w}||_2}\Big|.
\label{eq:app:two-time-alignemnt-update}
\end{equation}
This quantity measures the extent to which the eigenvector at time $t_w$ aligns with the parameter update accumulated over the subsequent $t$ steps. As seen in the figure, eigenvector at initial time seems to have the most lasting alignment with the update. Even though for $t=1$ there are significant alignments $\sim 0.1$, they do not persist in the weight change over larger $t$'s.

We also report here details of the displacement baseline computation.

\paragraph{Displacement Baseline}
To interpret what constitutes a meaningful difference between eigenvectors, we introduce a baseline defined by comparing top eigenvectors obtained from randomly sampled points in parameter space. Concretely, we fix the architecture and repeatedly initialize the model with different random seeds, computing the Hessian eigenvectors at each untrained initialization.

Under an idealized model in which eigenvectors are uniformly distributed on the unit sphere in $\mathbb{R}^N$, the expected cosine similarity between two independent vectors scales as $\mathcal{O}(1/\sqrt{N})$. More precisely, if the entries are assumed i.i.d., the resulting cosine similarity concentrates around zero with variance $\mathcal{O}(1/N)$. However, in practice these assumptions are violated: eigenvectors exhibit structured correlations and are typically more localized than Gaussian random vectors. As a result, the theoretical estimate is not quantitatively reliable, and an empirical baseline is more informative.

For a given architecture and dataset, we therefore estimate the baseline as the mean absolute cosine similarity between top eigenvectors extracted from 100 independently initialized networks. This yields $\tau = 0.0523$ (corresponding to $\tau^{(1)} = 2.1125 \times 10^{-5}$). For FashionMNIST, we obtain a slightly lower baseline of $\tau = 0.0492$ (i.e., $\tau^{(1)} = 2.1194 \times 10^{-5}$).

\paragraph{Connection with the cosine similarity metric}

From the displacement, we can directly recover the cosine similarity (i.e., the angle) between eigenvectors:
\begin{align*}
    \text{Cos-Sim}\Big(v^{(i)}(t_w), v^{(i)}(t_w + t)\Big) &= v^{(i)}(t_w)^\top v^{(i)}(t_w + t) \qquad \text{:  $v's$ are norm one}\\
    &= 1 - \frac{1}{2}(2 - 2v^{(i)}(t_w)^\top v^{(i)}(t_w + t) )\\
    & = 1 - \frac{1}{2}||v^{(i)}(t_w) - v^{(i)}(t_w + t)||_2^2
    = 1 - \frac{N}{2}\Delta^{(i)}(t_w, t_w+t).
\end{align*}
The absolute cosine similarity has been used in prior work to quantify eigenvector change due to its intuitive geometric interpretation as the angle between vectors \cite{bao_hessian_inertia_2023}. We nevertheless report displacement to maintain consistency with \cite{baity-jesi_comparing_2018}. Since absolute cosine similarity is linearly related to displacement, this choice does not affect the qualitative conclusions.

\clearpage

\section{Gradient Alignment - additional experiments}
\label{app:alignment}
We explicitly define the gradient and update alignment metrics used in Section \ref{sec:alignment}.Alignment of $i$th eigenvector after $t$ iterations $v^{(i)}(t)$, with direction of the gradient $\nabla \mathcal{L}_t$ at the same time is defined as
 \begin{equation}
     a^{(i)}_{\text{grad}}(t) = \Big|v^{(i)}(t)^T\frac{\nabla \mathcal{L}_t}{||\nabla \mathcal{L}_t||_2}\Big|.
     \label{eq:app:grad_vec_alignemnt}
 \end{equation}
Alignment of $i$th eigenvector after $t$ iterations $v^{(i)}(t)$, with parameter update $\theta_{t+1} - \theta_t$, which can be seen as how much of weight update is in the direction of top hessian eigenvector:
\begin{equation}
a^{(i)}_{\text{upd}}(t) = \Big|v^{(i)}(t)^T\Big(\frac{\theta_{t+1} - \theta_t}{||\theta_{t+1} - \theta_t||_2} \Big)\Big|.
\label{eq:app:update_alignemnt}
\end{equation}
Note that this differs from the two-time update alignment defined in Equation \ref{eq:app:two-time-alignemnt-update} in that it is defined with respect to consecutive steps rather than the waiting time $t_w$ and lag $t$. 

We next present empirical results for these metrics across training time and eigenvector rank.

\begin{figure}[h]
\centering
\begin{minipage}{0.8\textwidth}
    \centering
    \includegraphics[width=\linewidth]{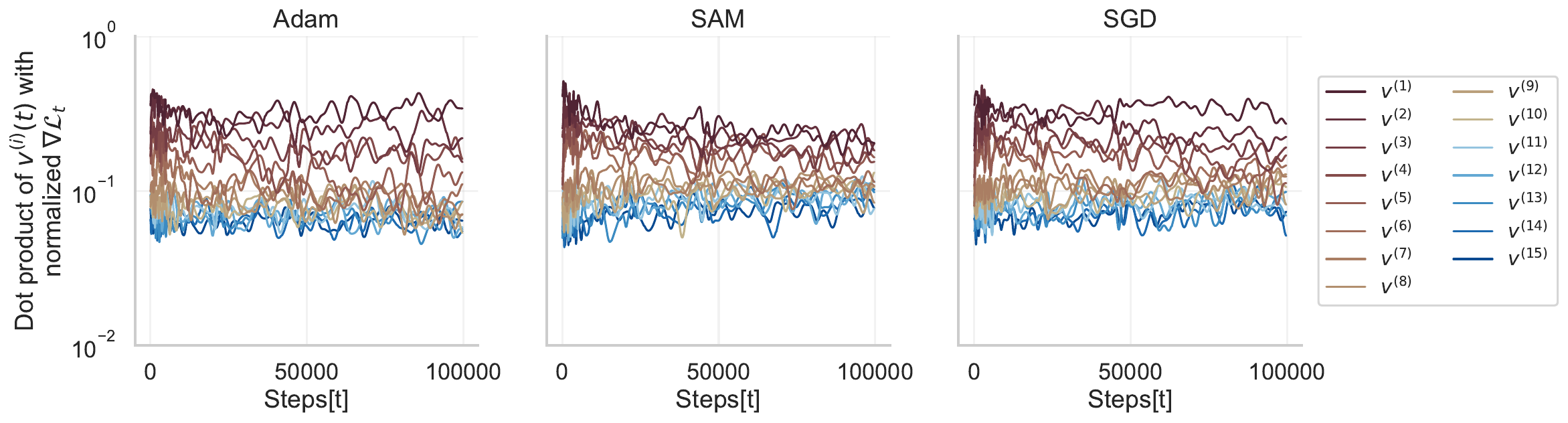}
    \caption*{(a) gradient alignment}
\end{minipage}
\hfill
\begin{minipage}{0.8\textwidth}
    \centering
    \includegraphics[width=\linewidth]{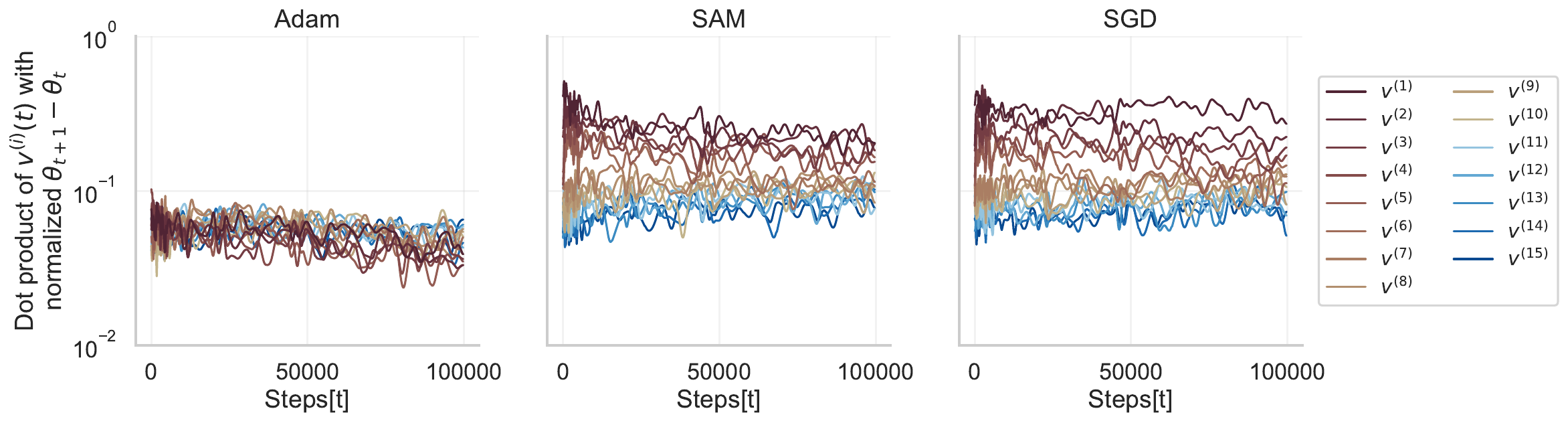}
    \caption*{(b) update alignemnt}
\end{minipage}

   \caption{Example of raw alignment of (normalized) gradient and update with the top curvature directions. This is FashionMNIST experiement. The values are gaussina smoothed with $\sigma=5$ and show no consistent trend, jsutifying use of mean aggregation as a more meaningful metric.}
    \label{fig:app:raw_alignemnt}
\end{figure}

Figure \ref{fig:app:raw_alignemnt} shows the raw (Gaussian-smoothed, $\sigma=5$) alignment of gradients and updates with the top eigenvectors over training. Despite smoothing, the curves do not exhibit a consistent temporal trend, with fluctuations dominating any systematic evolution. This motivates summarizing alignment via run-averaged statistics.

\begin{figure}[h]
\centering
\begin{minipage}{0.45\textwidth}
    \centering
    \includegraphics[width=\linewidth]{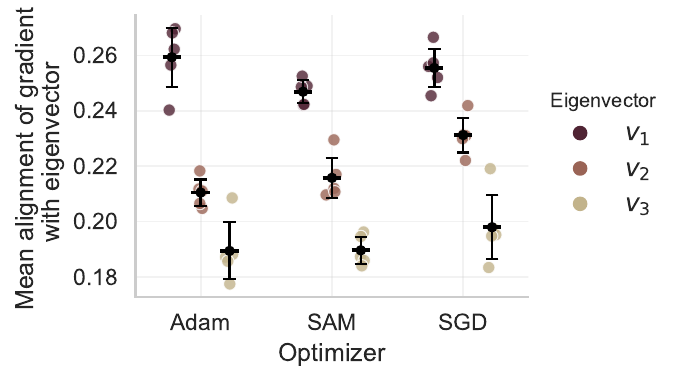}
    \caption*{(a) MNIST - gradient alignment}
\end{minipage}
\hfill
\begin{minipage}{0.45\textwidth}
    \centering
    \includegraphics[width=\linewidth]{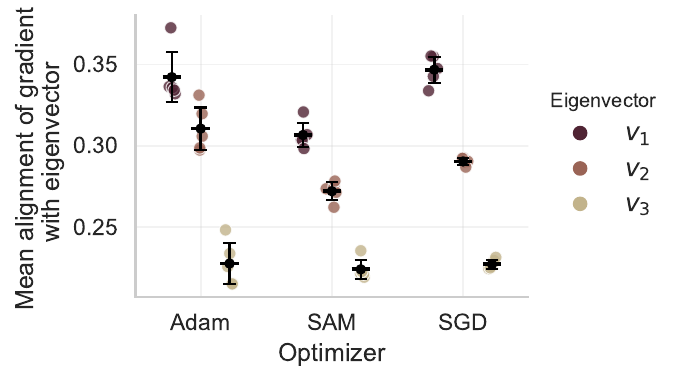}
    \caption*{(b) FashionMNIST - gradient alignment}
\end{minipage}
\begin{minipage}{0.45\textwidth}
    \centering
    \includegraphics[width=\linewidth]{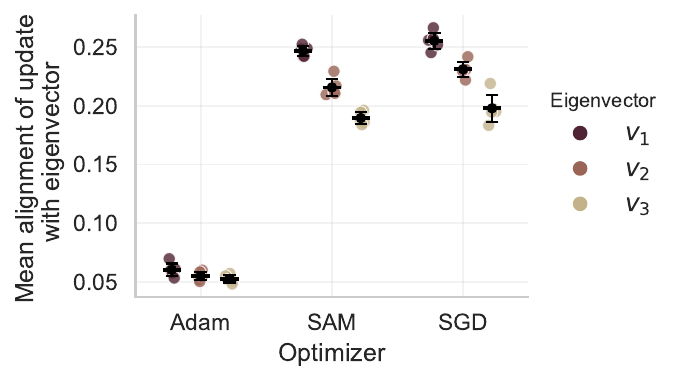}
    \caption*{(c) MNIST - update alignement}
\end{minipage}
\hfill
\begin{minipage}{0.45\textwidth}
    \centering
    \includegraphics[width=\linewidth]{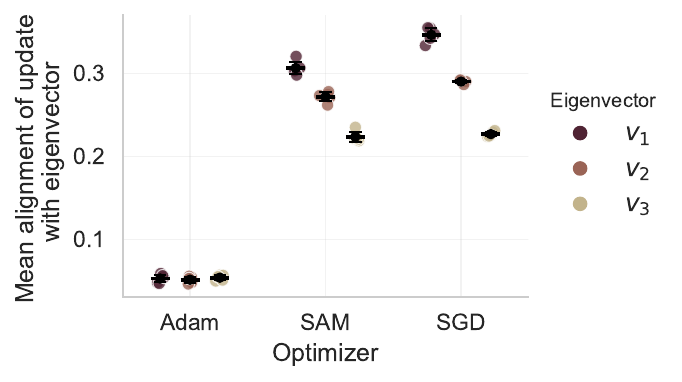}
    \caption*{(d) FashionMNIST - update alignemnt}
\end{minipage}

   \caption{Alignment of (normalized) gradient and update with the top curvature directions. Points in the plot correspond to different experiement runs.}
    \label{fig:app:alignment}
\end{figure}

Figure \ref{fig:app:alignment} reports these aggregated values across multiple initializations for MNIST and FashionMNIST. We observe some differences between optimizers: SGD exhibits the highest alignment with Hessian eigenvectors, SAM slightly reduces this alignment, and Adam remains significantly lower for update directions. 

\begin{figure}[h]
\centering
\begin{minipage}{0.75\textwidth}
    \centering
    \includegraphics[width=\linewidth]{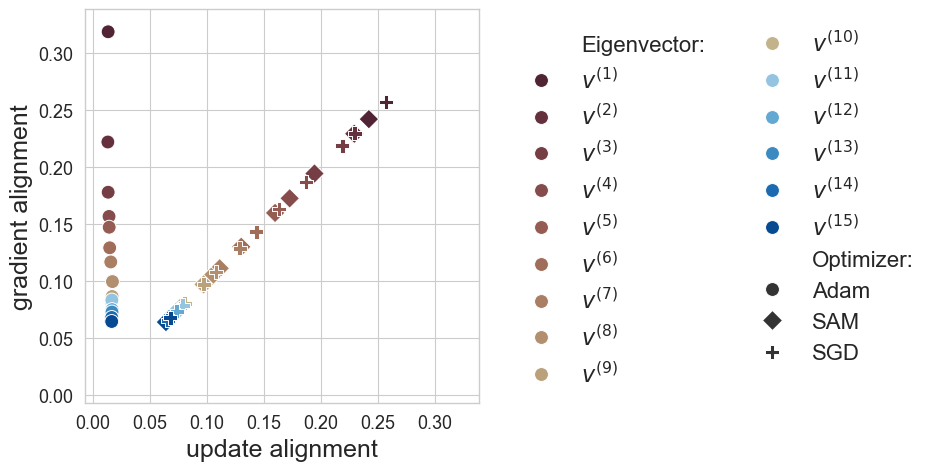}
    \caption*{(a) MNIST}
\end{minipage}
\hfill
\begin{minipage}{0.75\textwidth}
    \centering
    \includegraphics[width=\linewidth]{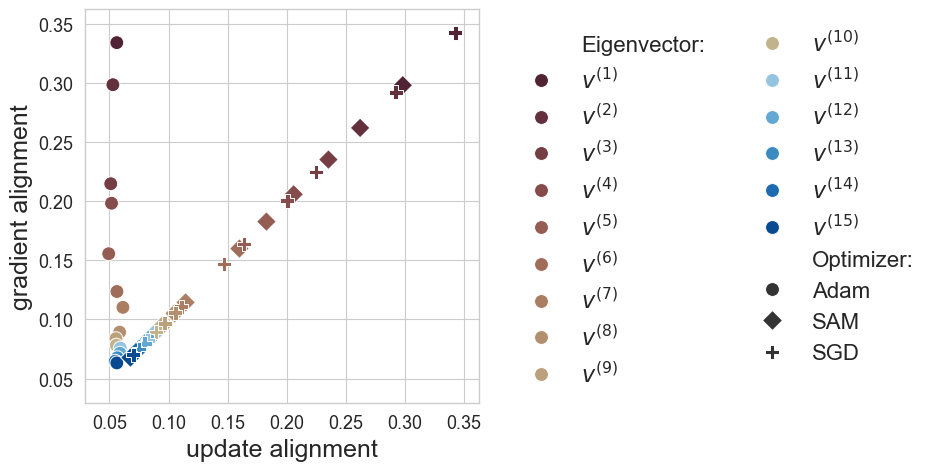}
    \caption*{(b) FashionMNIST}
\end{minipage}

   \caption{Alignment of (normalized) gradient and update with the top curvature directions. Points in the plot correspond to mean in a single run. We can observe that the value seems to depend on the eigenvalue magnitude.}
    \label{fig:app:15eigvs_alignemnt}
\end{figure}

Figure \ref{fig:app:15eigvs_alignemnt} further breaks this down across the top 15 eigenvectors. Here, a clear dependence on eigenvalue magnitude emerges: leading eigenvectors show substantially higher alignment. There is a mild transition around $i \geq 10$, beyond which alignment remains low. This confirms that only a small subset of dominant curvature directions strongly interact with both gradients and updates.

\paragraph{Gradient alignment}
It has been observed that for gradient descent, the gradient aligns with each of the top-$k$ Hessian eigenvectors with magnitude approximately $1/k$, where $k$ is the number of classes \cite{gur-ari_gradient_2018}. In contrast, we observe a dependence on eigenvalue magnitude, with the mean $a_{\text{grad}}^{(1)}$ significantly above $1/k = 0.1$. This discrepancy is likely due to the use of stochastic mini-batch gradients, where the gradient is computed on subsets of the data, whereas the Hessian is defined over the full dataset. Nevertheless, assuming that gradients remain primarily contained within the top Hessian subspace, this provides a basis for relating update–eigenvector alignment to the degree of alignment between the optimizer’s update direction and the gradient.

\paragraph{Update alignment}
As expected, SGD update directions exhibit the strongest alignment with the Hessian eigenvectors, while SAM modifies this behaviour and Adam remains around $0.05$. For SGD, the update direction is directly given by the gradient, so $a_{\text{upd}}^{(i)} = a_{\text{grad}}^{(i)}$, and the results on gradient–eigenvector alignment \cite{gur-ari_gradient_2018} directly carry over to update alignment.

For Adam, the update rule is given by
\[
m_t = \beta_1 m_{t-1} + (1-\beta_1)\nabla_{\theta_t}\mathcal{L},
\]
where $m_t$ denotes the momentum-based update direction. With $\beta_1 = 0.9$ used in our training, the gradient contributes a weight of $1-\beta_1 = 0.1$, in addition to the momentum term $m_{t-1}$. This yields the lower bound
\[
a_{\text{upd}}^{(i)}(t) \geq (1-\beta_1)\, a_{\text{grad}}^{(i)}(t).
\]
In the MNIST experiment, we observe $\mathbb{E}_t[a_{\text{grad}}^{(i)}(t)] \approx 0.26$ (see Figure \ref{fig:app:alignment}), which implies a contribution of at least $0.026$ from the gradient component alone. The observed mean $a_{\text{upd}}^{(1)}$ of approximately $0.05$ further suggests that empirically
\[
\mathbb{E}\!\left[(m_{t-1})^T v^{(i)}(t)\right] \approx \frac{0.05 - 0.026}{0.9} \approx 0.024,
\]
indicating that the momentum term itself exhibits relatively low alignment with the Hessian eigenvectors. Similar observations have been reported in prior work \cite{song_does_2025}.

The SAM update incorporates second-order information more explicitly. The effective update direction can be approximated as
\[
\nabla_{\theta} \mathcal{L} + \rho H \frac{\nabla_{\theta} \mathcal{L}}{\|\nabla_{\theta} \mathcal{L}\|_2},
\]
with $\rho = 0.05$ in our experiments. This second-order correction can slightly reduce following the gradient relative to SGD. However, as shown in Figure \ref{fig:app:alignment}, we still observe that $a_{\text{upd}}^{(i)} \approx a_{\text{grad}}^{(i)}$, suggesting that the slightly smaller update alignment is likely attributable to a flatter loss landscape rather than the update rule itself.

\clearpage

\section{Localization - additional experiments}
\label{app:loclaization}
We present a few more figures complementary to the section's \ref{sec:IPR} results.

\begin{figure}
\begin{minipage}{\textwidth}
    \centering
    \includegraphics[width=\linewidth]{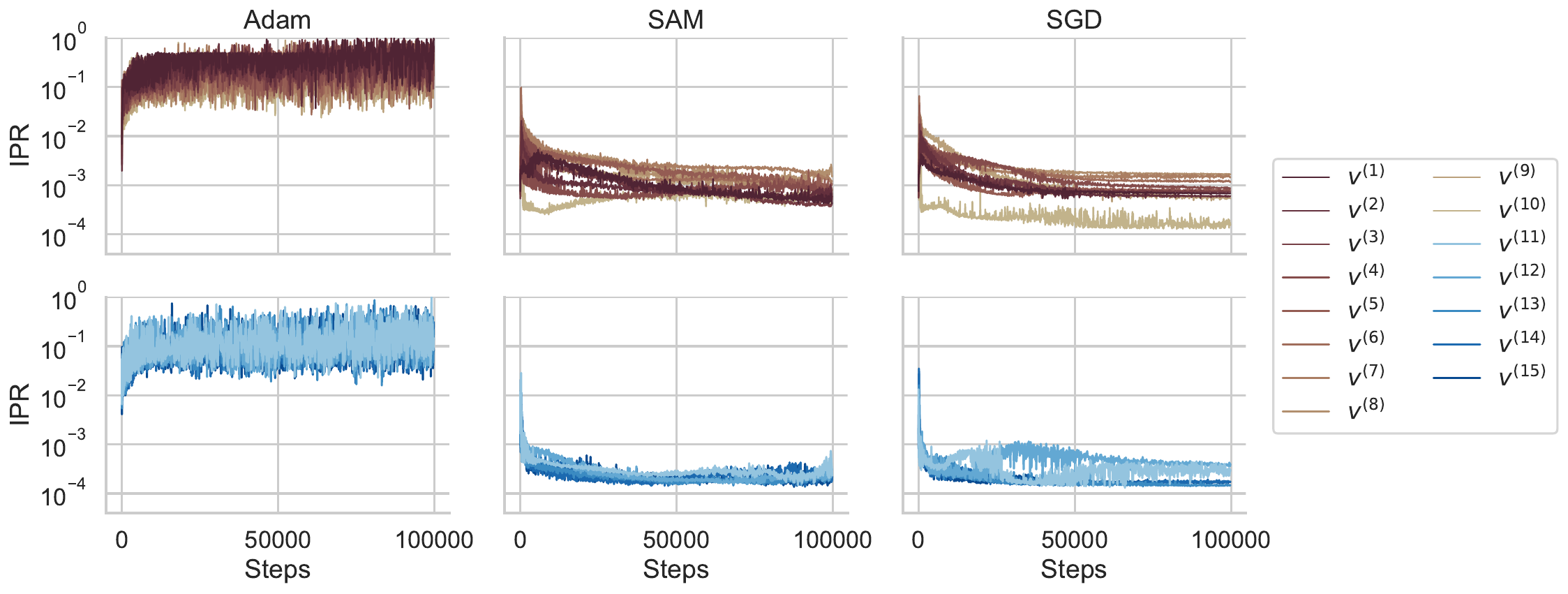}
    \caption*{(a) MNIST}
\end{minipage}
\hfill
\begin{minipage}{\textwidth}
    \centering
    \includegraphics[width=\linewidth]{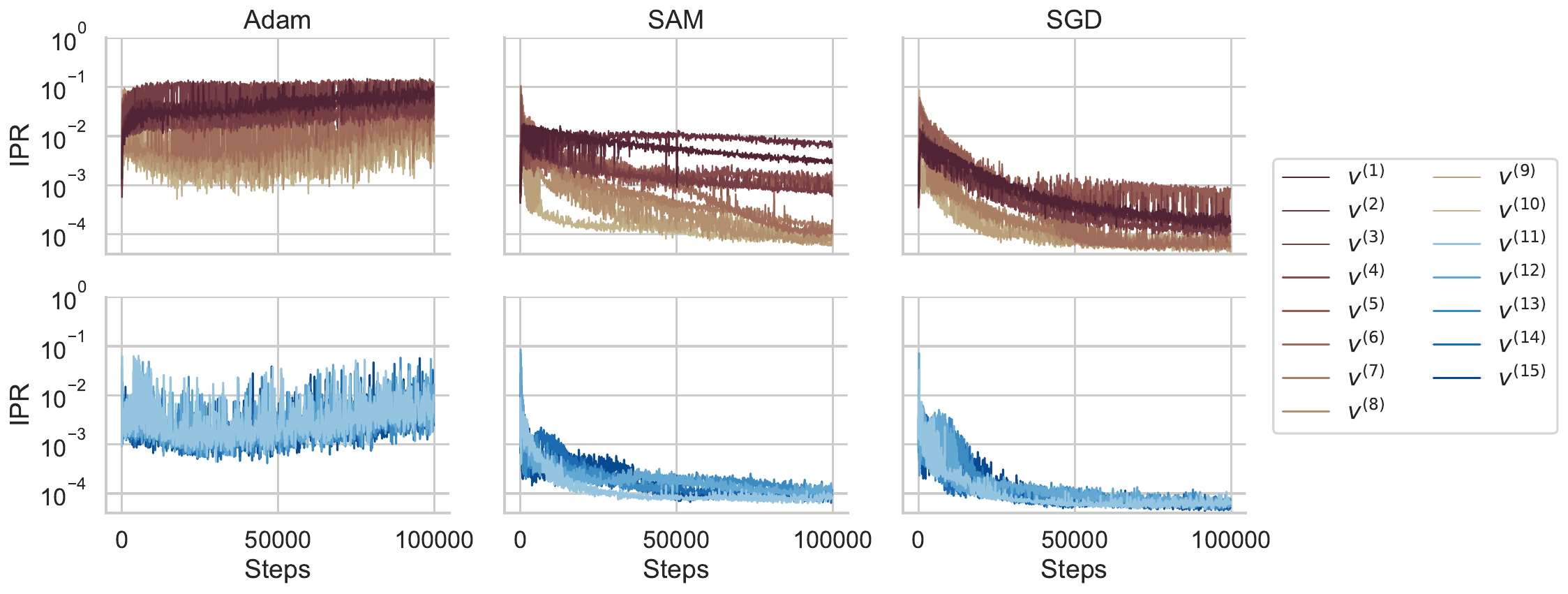}
    \caption*{(b) FashionMNIST}
\end{minipage}
    \caption{Inverse participation ratio (IPR) of the top 15 eigenvectors throughout training for a single run. Top plot presents eigenvectors $v^{(1)}-v^{(10)}$, and the bottom $v^{(11)}-v^{(15)}$  for visual clarity. The consistent grouping of neighbouring trajectories suggests that consecutive eigenvectors exhibit similar localization behaviour.}
    \label{fig:app:15eig_IPR}
\end{figure}

\paragraph{Localization patterns extend over the leading eigenvector}
In section \ref{sec:IPR} we observe that the Inverse Participation Ratio (IPR) is a statistic of the eigenvector that admits a stable trend. 
We observe that even with strong eigenvector mixing, IPR seems to be a statistic that doesn't vary strongly. Hence, we expect that other top eigenvectors also follow the same localization values. Indeed, in Figure \ref{fig:app:15eig_IPR} we plot the IPR of the top 15 eigenvectors evaluated over a single run and observe that they share the similar IPR values. This is particularly visible for the MNIST training - with a seeming separation at $v^{(9)}$ and $v^{(10)}$. Many of the observations reported in section \ref{sec:IPR} still hold:
\begin{itemize}
    \item Adam exhibits a localization trend, with IPR values converging toward $\mathcal{O}(10^{-1})$, whereas SGD remains strongly delocalized, stabilizing around $\mathcal{O}(10^{-4})$.
    
    \item This persistent difference in localization across eigenvectors confirms that the different optimizers operate in qualitatively distinct regions of the loss landscape.
    
    \item The overall pattern is consistent across both training on MNIST and FashionMNIST, indicating that these effects are not dataset-specific.
    
    \item SGD and SAM display surprisingly low step-to-step variance; the raw IPR trajectories are notably smooth despite stochastic training.
\end{itemize}

\begin{figure}
\begin{minipage}{0.45\textwidth}
    \centering
    \includegraphics[width=\linewidth]{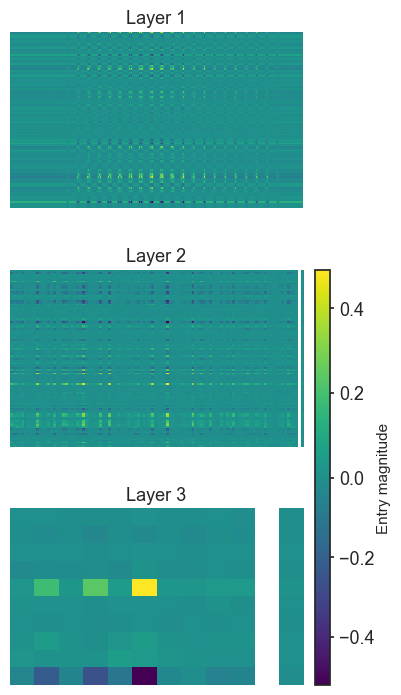}
    \caption*{(a) Adam $v^{(i)}(t_{\text{final}})$}
\end{minipage}
\hfill
\begin{minipage}{0.45\textwidth}
    \centering
    \includegraphics[width=\linewidth]{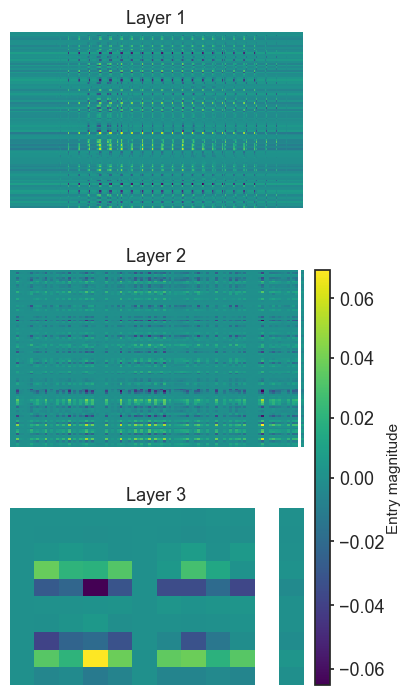}
    \caption*{(b) SGD $v^{(i)}(t_{\text{final}})$}
\end{minipage}
       \caption{Comparison of the reshaped leading eigenvector at the end of training on MNIST. The entries of the eigenvector are mapped into the shape of corresponding network parameters (weights and biases shown adjacently). A notable difference in scale is observed: the colour range for Adam is approximately seven times larger than that of SGD. This indicates that the leading eigenvector under Adam is more localized, with mass concentrated in a small subset of parameters, whereas under SGD it is more diffusely distributed across many entries.
}
    \label{fig:app:reshaped}
\end{figure}

\paragraph{Visualizing the eigenvectors} Since the eigenvector entries directly correspond to model parameters, we can plot the eigenvector entries in the shape of model weights. Figure \ref{fig:app:reshaped} shows the leading eigenvector at the end of Adam MNIST training - it is visibly localized with many entries having values $\sim 0$. This visualization not only remind us of the spatial meaning but also connects it to the network's gradient propagation. Hessian-vector-product (that we base the eigenvector estimation on) is evaluated by two back-propagations through the network. This is reflected in the localization - the activated entries in layer 3 have corresponding activated entries in layer 2 of the same sign.


\clearpage
\section{Impact of batch size and learning rate on the results}
\label{app:ablation}

Batch size and learning rate are known to affect Hessian dynamics during training. Large-batch training typically explores sharper regions of the loss landscape \cite{yao2018hessian}, while increasing the learning rate can induce instabilities that lead to rotations of the leading Hessian eigenspaces and favour flatter solutions \cite{WANG2026108874}. Since both effects can modify the Hessian spectrum and promote eigenvector mixing, we investigate whether the localization, displacement, and alignment phenomena observed in the main text remain robust under variations of batch size and learning rate.

\subsection{Implementation details}

In the batch-size ablation, we use a wide range of
$$\text{Batch sizes: }\{32, 64, 128, 1024, 60000\},$$
including 128 used in the main experiment, as well as full-batch update using all 60000 samples.

To report comparable results between runs, the number of update steps for batch size $B$ is adjusted as follows $N_{B} = \lfloor 10^5\cdot \frac{128}{B} \rfloor$. This adjustment ensures that all runs process the same amount of information. 

In the learning rate ablation, we train using five learning rates centred around the one considered in the main experieemnts. That is
$$\text{Adam Learning Rates: }\{10^{-2}, 3\cdot 10^{-3}, 10^{-3}, 3\cdot 10^{-4}, 10^{-4}\}$$
$$\text{SGD Learning Rates: }\{10^{-1}, 3\cdot 10^{-2}, 10^{-2}, 3\cdot 10^{-3}, 10^{-3}\}$$
All runs still evaluate eigenvectors at 1000 different time steps, with the times of evaluation being adjusted over the total number of update iterations $N_B$. We restrict the experiements to Adam and SGD, keeping all the optimzier parameters unchanged. Each setup is run over 3 random seeds.
We keep all other parameters and setting fixed, to fit the original setup.


\subsection{Batch size impact results}
\begin{figure}
\begin{minipage}{\textwidth}
    \centering
    \includegraphics[width=0.8\linewidth]{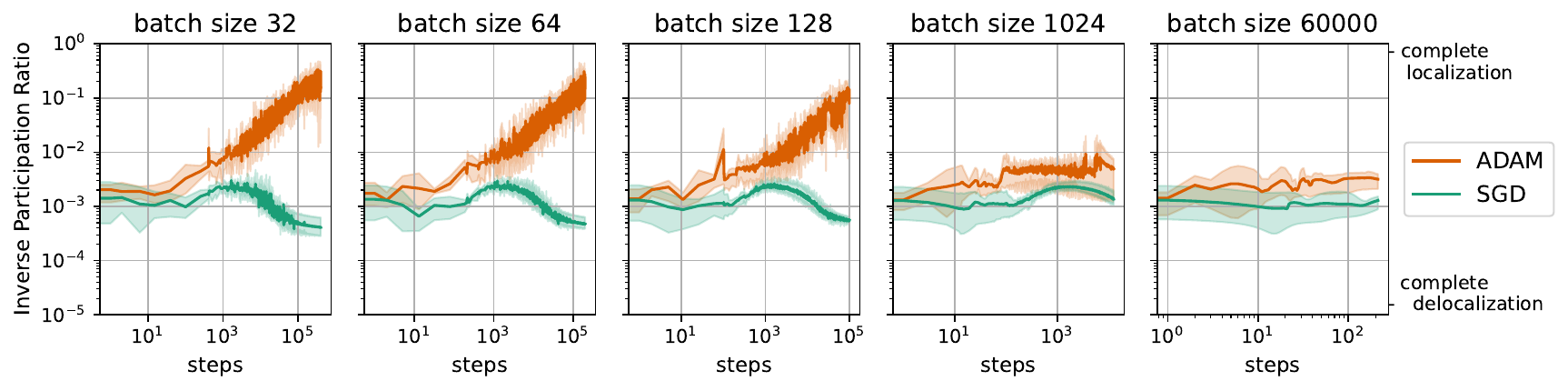}
    \caption*{(a) MNIST}
\end{minipage}
\hfill
\begin{minipage}{\textwidth}
    \centering
    \includegraphics[width=0.8\linewidth]{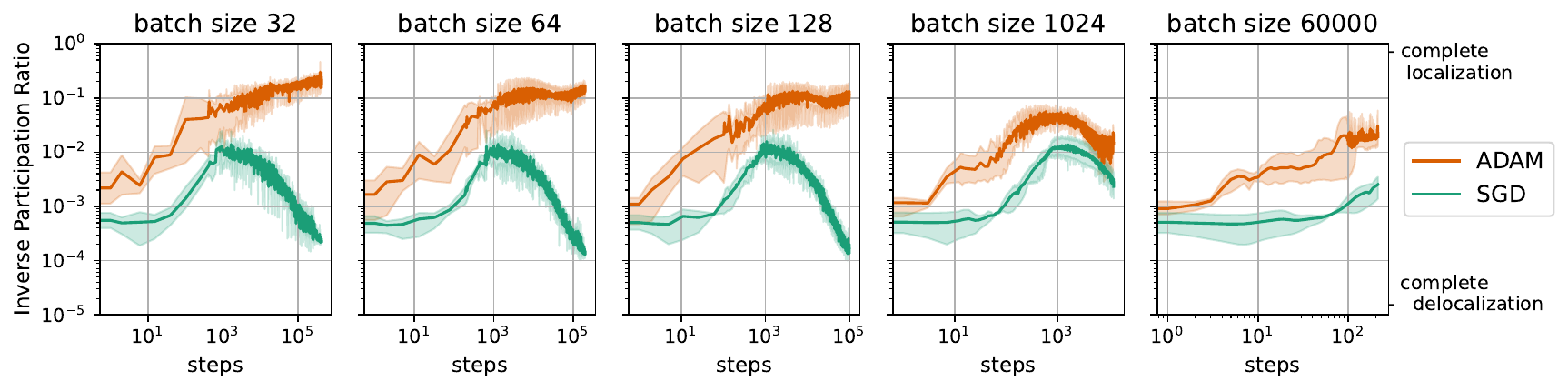}
    \caption*{(b) FashionMNIST}
\end{minipage}
    \caption{Inverse participation ratio (IPR) of the leading eigenvector throughout training for varying batch sizes. Mean and min-max over 3 initializations. The training using Adam progressively localizes the eigenvectors while SGD-based optimizers have a delocalizing trend for small to moderate-size batches, with a very small variation between runs.}
    \label{fig:app:abltion_batch_IPR}
\end{figure}

\begin{figure}
    \centering
    \includegraphics[width=1\linewidth]{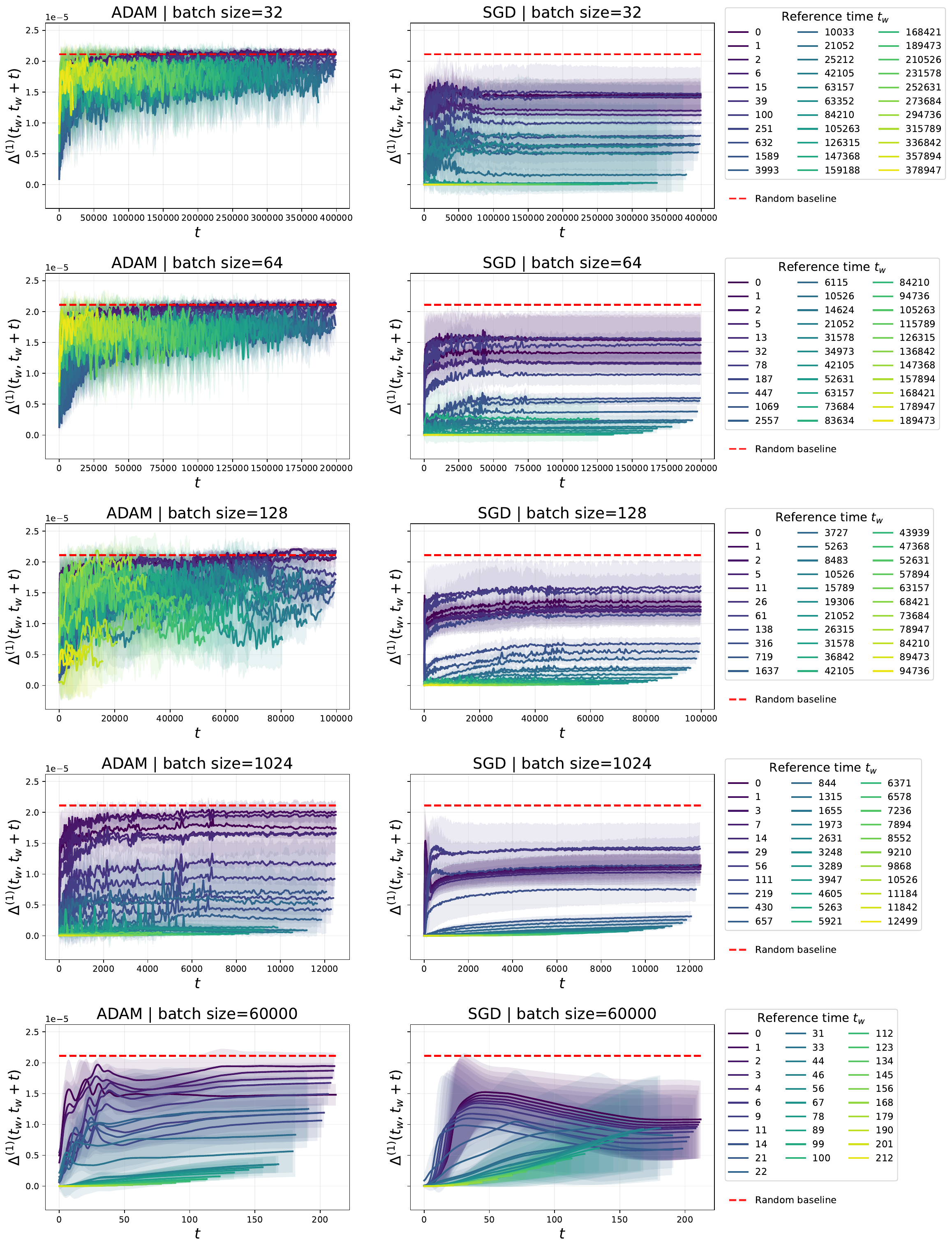}
    \caption{Two-time mean square displacement, $\Delta^{(i)}(t_w, t_w + t)$ as defined in Equation \ref{eq:displacement}. Each curve
$t_w$ corresponds to the leading eigenvector at time $t_w$ and its diffusion after waiting $t$ steps. Number of training steps adjusted to be comparable between differnt batch size runs. Red dashed line indicates the random baseline. Observations from the main text still hold, with the exception of full-batch setting (batch size=60000).}
    \label{fig:app:abla-batch:displacement}
\end{figure}

\begin{itemize}
\item Figure \ref{fig:app:ablation_alignment} presents the \textbf{alignment for varying batch size}.  There is a smooth trend between batch sizes with a sharp transition for full batch training.
\item Figure \ref{fig:app:abltion_batch_IPR} presents the \textbf{IPR trajectories for varying batch sizes}. The optimizer-dependent localization patterns remain largely unchanged across batch sizes, except for very large and full-batch regimes. As larger batches perform fewer parameter updates over the same number of epochs, their trajectories resemble earlier stages of the corresponding smaller-batch runs.

\item Figure \ref{fig:app:abla-batch:displacement} shows the \textbf{displacement trajectories for varying batch sizes}. SGD remains below the random baseline for all but full-batch training, while Adam diffuses more rapidly and typically reaches the baseline. The qualitative difference in eigenvector diffusion between the optimizers persists across batch sizes, although it is weaker on FashionMNIST.
    
\end{itemize}

The full-batch regime should be interpreted separately. While all experiments use the full-data Hessian, only the mini-batch setting introduces gradient noise; full-batch training additionally exhibits distinct Hessian dynamics associated with the Edge of Stability.

\subsection{Learning rate impact results}

\begin{figure}
    \centering
    \includegraphics[width=0.85\linewidth]{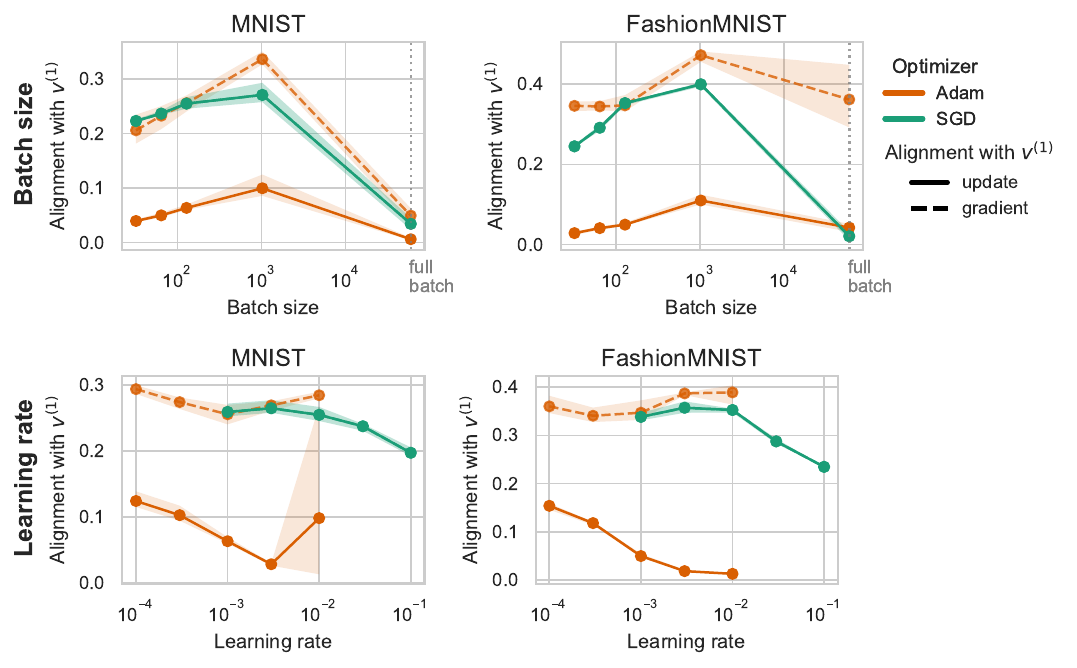}
    \caption{Dot product of normalized gradient and update with $v^{(1)}$ for, as defined in Equations \ref{eq:app:update_alignemnt}, \ref{eq:app:grad_vec_alignemnt}. Points in the plot correspond to mean alignment values over different experiment initializations, shaded areas to min-max over 3 runs. Gradient and update alignment naturally overlaps for SGD. There is a sharp difference with the full-batch training.}
\label{fig:app:ablation_alignment}
\end{figure}

\begin{figure}
\begin{minipage}{\textwidth}
    \centering
    \includegraphics[width=0.8\linewidth]{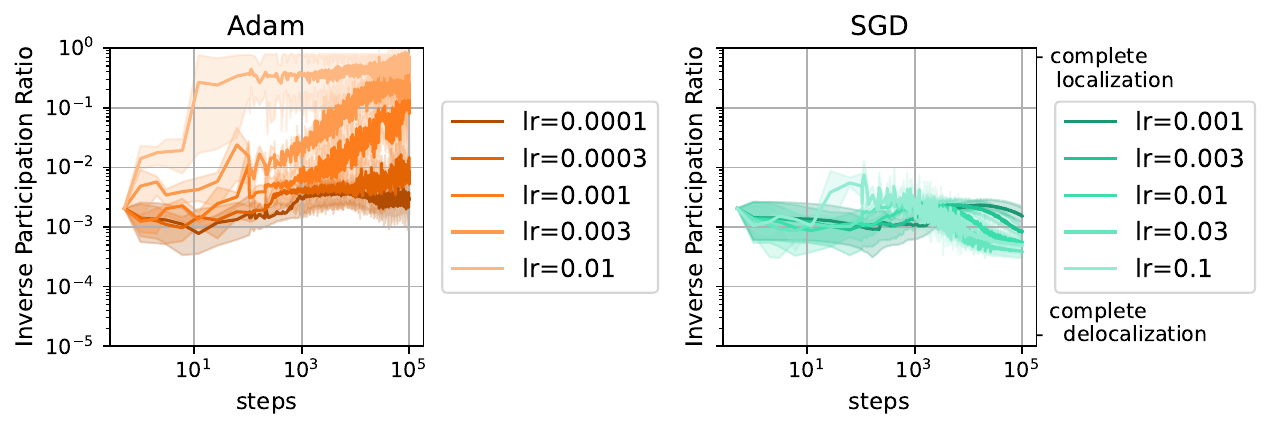}
    \caption*{(a) MNIST}
\end{minipage}
\hfill
\begin{minipage}{\textwidth}
    \centering
    \includegraphics[width=0.8\linewidth]{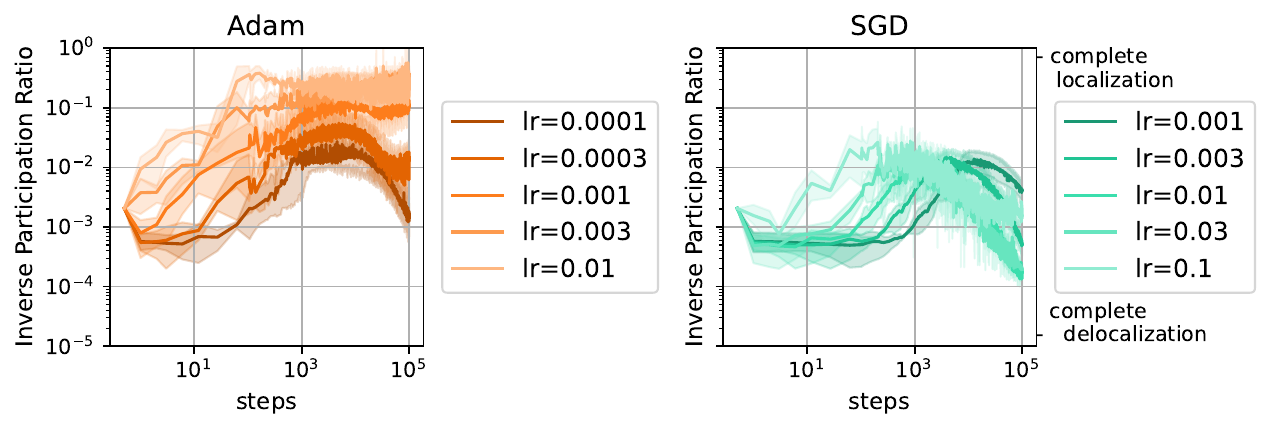}
    \caption*{(b) FashionMNIST}
\end{minipage}
    \caption{Inverse participation ratio (IPR) of the leading eigenvector throughout training for varying learning rates. Mean and min-max over 3 initializations.}
    \label{fig:app:learning_rate_ablation_IPR}
\end{figure}

\begin{figure}
    \centering
    \includegraphics[width=0.9\linewidth]{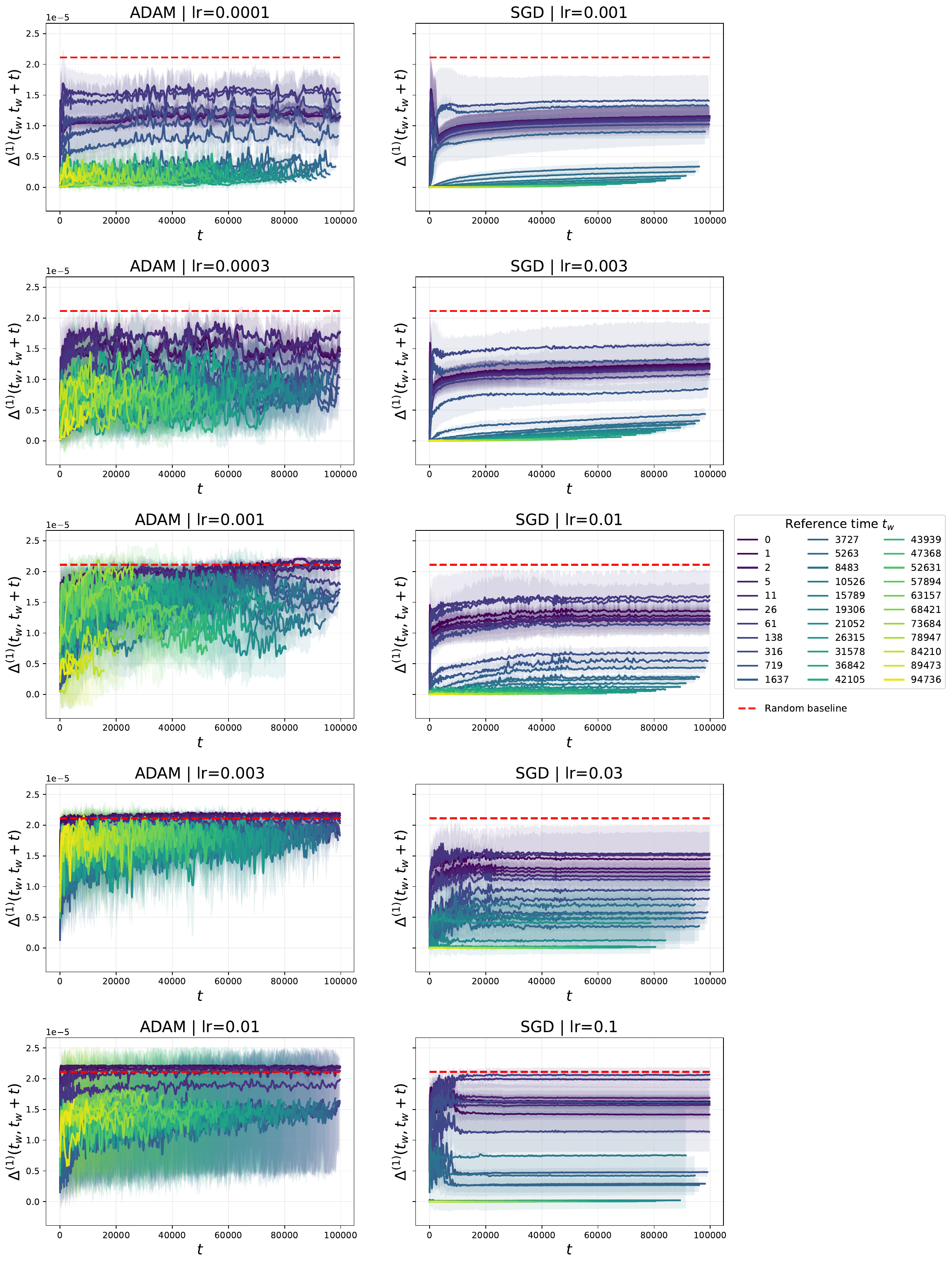}
    \caption{Two-time mean square displacement, $\Delta^{(i)}(t_w, t_w + t)$ as defined in Equation \ref{eq:displacement}. Each curve
$t_w$ corresponds to the leading eigenvector at time $t_w$ and its diffusion after waiting $t$ steps. Red dashed line indicates the random baseline. The learning rate doesn't impact the differences in eigenvector displacement across optimizers.}
    \label{fig:app:displacement_learning_rate_ablation}
\end{figure}

\begin{itemize}
    
    \item Figure \ref{fig:app:ablation_alignment} presents the \textbf{alignment across varying learning rates}. We observe the expected behavior: Adam maintains low update alignment across all learning rates, while gradient alignment is comparable to SGD. 
    \item Figure \ref{fig:app:learning_rate_ablation_IPR} presents the \textbf{IPR across varying learning rates}. For Adam, increasing the learning rate leads to rapid localization within approximately 10 steps, while this effect does not hold at small learning rates. SGD remains largely invariant across learning rates, with near-neutral values and slightly higher variability on FashionMNIST.
    \item Figure \ref{fig:app:displacement_learning_rate_ablation} presents the \textbf{displacement trajectories for varying learning rates}. The qualitative distinction between Adam and SGD persists across learning rates. Larger learning rates accelerate displacement, with Adam reaching the random baseline for sufficiently large values. SGD eigenvectors appear largely converged and continue to exhibit aging behavior.
\end{itemize}

\end{document}